\documentclass[conference]{IEEEtran}
\usepackage{cite}
\usepackage{amsmath}
\usepackage{amssymb,amsfonts}
\usepackage{algorithm}
\usepackage{algorithmic}
\usepackage{graphicx}
\usepackage{textcomp}
\usepackage{xcolor}
\def\BibTeX{{\rm B\kern-.05em{\sc i\kern-.025em b}\kern-.08em
    T\kern-.1667em\lower.7ex\hbox{E}\kern-.125emX}}

\usepackage{multirow,multicol}
\usepackage{booktabs}

\usepackage{hyperref}

\usepackage{cleveref}
\crefname{equation}{Equation}{Equations}
\Crefname{equation}{Equation}{Equations}
\crefname{table}{Table}{Tables}
\Crefname{table}{Table}{Tables}
\crefname{figure}{Fig.}{Figs.}
\Crefname{figure}{Fig.}{Figs.}
\crefname{section}{Section}{Sections}
\Crefname{section}{Section}{Sections}
\crefname{algorithm}{Algorithm}{Algorithms}
\Crefname{algorithm}{Algorithm}{Algorithms}

\def\proposedmethod{TEMP}
\def\eg{\textit{e.g.}}
\def\ie{\textit{i.e.}}
\def\etal{\textit{et al.}}
\def\vs{\textit{vs.}}

\newcommand{\inputfig}[6][]{
\begin{figure#1}[tb]
\centering
\includegraphics[alt={#6},width=#2\linewidth]{#3}
\caption{#4}
\label{#5}
\end{figure#1}
}

\begin{document}

\title{Test-time Similarity Modification \\for Person Re-identification\\toward Temporal Distribution Shift
}

\author{\IEEEauthorblockN{Kazuki Adachi$^*${\quad}Shohei Enomoto$^*${\quad}Taku Sasaki$^*${\quad}Shin'ya Yamaguchi$^{*\dagger}$}
\IEEEauthorblockA{$^*$NTT Corporation, Tokyo, Japan \\
$^\dagger$Kyoto University, Kyoto, Japan \\
\{kazuki.adachi, shohei.enomoto, taku.sasaki, shinya.yamaguchi\}@ntt.com}
}

\maketitle

\begin{abstract}
Person re-identification (re-id), which aims to retrieve images of the same person in a given image from a database, is one of the most practical image recognition applications.
In the real world, however, the environments that the images are taken from change over time.
This causes a distribution shift between training and testing and degrades the performance of re-id.
To maintain re-id performance, models should continue adapting to the test environment's temporal changes.
Test-time adaptation (TTA), which aims to adapt models to the test environment with only unlabeled test data, is a promising way to handle this problem because TTA can adapt models instantly in the test environment.
However, the previous TTA methods are designed for classification and cannot be directly applied to re-id.
This is because the set of people's identities in the dataset differs between training and testing in re-id, whereas the set of classes is fixed in the current TTA methods designed for classification.
To improve re-id performance in changing test environments, we propose TEst-time similarity Modification for Person re-identification ({\proposedmethod}), a novel TTA method for re-id.
{\proposedmethod} is the first fully TTA method for re-id, which does not require any modification to pre-training.
Inspired by TTA methods that refine the prediction uncertainty in classification, we aim to refine the uncertainty in re-id.
However, the uncertainty cannot be computed in the same way as classification in re-id since it is an open-set task, which does not share person labels between training and testing.
Hence, we propose re-id entropy, an
 alternative uncertainty measure for re-id computed based on the similarity between the feature vectors.
Experiments show that the re-id entropy can measure the uncertainty on re-id and {\proposedmethod} improves the performance of re-id in online settings where the distribution changes over time.
\end{abstract}


\section{Introduction} \label{sec:introduction}
Image recognition based on deep neural networks has achieved high accuracy~\cite{alexnet,Simonyan2015VeryDC,resnet,dosovitskiy2021an}.
In particular, person re-identification (re-id)~\cite{Ahmed_2015_CVPR}, which aims to retrieve images of persons from a database (gallery) that have the same identity as a given image (query), is one of the most practical applications of image recognition.

In the real world, test environments, \eg, cities or buildings, differ from the training environment and further change temporally due to location changes and image corruption.
Such changes cause a distribution shift and degrade the performance of re-id~\cite{Deng_2018_CVPR,Wang_2018_CVPR}.
In such a situation, re-id models should adapt to the distribution shift to retain accuracy.
One possible way is to re-train the models on data collected from the test environment by fine-tuning or unsupervised domain adaptation (UDA)~\cite{re-id_uda_survey}.
However, collecting and storing the data can be prohibitive in re-id because images of persons may have privacy or security concerns.
From this perspective, images should be discarded immediately after prediction rather than stored and used for re-training.

To find a way to maintain re-id models' accuracy against changes in test environments, this paper investigates an adaptation method that works in an online manner, \ie, each mini-batch from the test environment (target domain) is used for model updates and is predicted sequentially.
An online adaptation method would resolve the above concerns since each mini-batch can be discarded immediately after prediction.
This problem can be formulated as \emph{test-time adaptation (TTA)}.
TTA~\cite{Wang2021,zhou2021bayesian,niu2022efficient} is a problem setting that has been actively studied to adapt a given source-pretrained model to incoming target data instantly without accessing the source dataset.
However, existing TTA methods are designed for closed-set classification, where the class set is fixed and shared between training and testing.
These TTA methods for classification cannot be naively applied to open-set recognition like re-id, where the set of person identities in the dataset differs between the training and testing.
Since TTA for open-set recognition is still an open problem, we need to newly develop a TTA methodology for re-id.
Specifically, we focus on \textit{Fully} TTA (\textit{FTTA}), which does not require any special modification~\cite{han2022generalizable} to the architecture or source-pretraining procedure~\cite{zhou2021bayesian}.
FTTA is applicable to broader cases than TTA because FTTA can reuse and adapt arbitrary off-the-shelf models pre-trained in arbitrary ways.

This paper proposes \emph{\textbf{te}st-time similarity \textbf{m}odification for \textbf{p}erson re-identification ({\proposedmethod})}, a novel and simple FTTA method for re-id models.
Inspired by a representative TTA approach for classification that minimizes uncertainty of model outputs~\cite{Wang2021,zhou2021bayesian,niu2022efficient}, we seek to minimize uncertainty of predictions for re-id models.
However, unlike classification models, the outputs of re-id models are feature vectors.
Since re-id is an open-set task, re-id predictions are based on the similarities of the output feature vectors instead of computing logits or probabilities for a fixed set of classes.
This gap between re-id and classification models hinders computing prediction uncertainty (\eg, entropy) naively as classification.
Thus, we propose \textit{re-id entropy}, an uncertainty measure for open-set recognition.
Re-id entropy is computed based on the similarities between the query and gallery features.
Although in classification, it is known that entropy strongly correlates with the accuracy~\cite{Wang2021}, whether the same relationship holds between re-id entropy and the performance of re-id is non-trivial since they are computed in different ways from those of classification.
But we experimentally found that re-id entropy also correlates with the performance of re-id and minimizing re-id entropy improves the performance of re-id.


We carried out experiments on Market-1501~\cite{Zheng_2015_ICCV}, MSMT17~\cite{Wei_2018_CVPR}, and PersonX~\cite{Sun_2019_CVPR}, which are representative re-id datasets.
The results showed that {\proposedmethod} improves the performance of re-id by up to about nine points compared with the TTA baselines under temporal distribution shifts in a batched online manner without accessing source data.

Our contributions in this paper are summarized as the followings:
\begin{itemize}
\item We propose {\proposedmethod}, a novel FTTA method for re-id.
{\proposedmethod} is the first FTTA approach for re-id that enables to reuse and adapt arbitrary off-the-shelf models trained in arbitrary ways during testing.
\item To adapt re-id models, we propose re-id entropy, a counterpart of entropy in classification that measures uncertainty of re-id prediction.
We experimentally confirmed that re-id entropy strongly correlates with the performance of re-id.
\item We show that {\proposedmethod} outperforms the baselines in re-id tasks under temporal distribution shifts such as location change and image corruption.
\end{itemize}
\section{Related Work} \label{sec:related_work}
\subsection{Unsupervised Domain Adaptation for Person Re-identification}
We refer to unsupervised domain adaptation (UDA) for re-id, not a general UDA for classification.
UDA is an actively studied field in re-id for transferring source domain knowledge to the target domain~\cite{re-id_uda_survey}.
UDA does not require labels for the target data and thus can avoid the high annotation costs of supervised re-training in the target domain.
UDA methods are mainly divided into two approaches: refining features with pseudo-labels or absorbing the feature differences in camera conditions.
The former approach uses clustering~\cite{lin2019bottom,NEURIPS2020_821fa74b} and/or a mean teacher model~\cite{ge2020mutual} for pseudo-labeling.
The latter approach uses adversarial training~\cite{zhang2021unsupervised} or style transfer~\cite{Zhong_2019_CVPR} for leveraging camera invariance.
These methods work well when adapting models to other domains different to the source domain, such as cameras or locations.
They assume that the target domain is stable and its distribution does not change.
They adapt in an offline manner, \ie, the whole target dataset is stored and used for multiple epochs for clustering, pseudo-labeling, and re-training.
However, target domains in the real world are unstable~\cite{niu2022efficient}, and adaptation in an offline manner that stores and reuses old data cannot follow the temporal changes.
Furthermore, the collected data may raise privacy concerns.
As a way of continually adapting to the changing domain while maintaining the old knowledge, CLUDA-ReID~\cite{Huang_2022_CVPR} introduces a meta-learning-based approach that balances the knowledge of the past and current domains.
However, CLUDA-ReID performs clustering for pseudo-labeling with holding the past data for every few epochs, which can be sensitive and makes adaptation done in an offline manner.

In contrast to these prior studies, we seek to adapt re-id models to the unstable target domains in an online manner.

\subsection{Test-time Adaptation}
In classification, test-time adaptation (TTA) has recently been studied to adapt a source-pretrained model to the target domain with only an unlabeled target dataset.
The difference between TTA and UDA is that TTA can be applied in an online manner, \ie, each mini-batch of the target data is used for updating the model and is predicted sequentially.
Since each mini-batch can be discarded immediately after the prediction, TTA can alleviate storage and privacy concerns.
One simple way is to update the statistics of the batch normalization (BN) layers~\cite{batch-norm} with incoming mini-batches at test time (BN-adapt~\cite{Benz_2021_WACV}), which can adapt to simple distribution shifts in classification.
Tent~\cite{Wang2021}, which is a representative method of TTA, refines the confidence of predictions by minimizing the entropy of the prediction probability on the target dataset because Wang \etal~\cite{Wang2021} found that the entropy strongly correlates with accuracy.
DELTA~\cite{zhao2023delta} modifies update of BN statistics and introduces dynamic weighting of loss to alleviate class-imbalance.
These TTA methods mentioned above are especially called \textit{Fully} TTA (FTTA), which does not require any special modifications on the architecture or source-pretraining procedure, whereas the following TTA methods requires some modifications.
BACS~\cite{zhou2021bayesian} uses maximum-a-posteriori regularization for the model parameters in addition to entropy minimization.
EATA~\cite{niu2022efficient} further addresses adaptations to continuously changing target domains.

Although these TTA methods resolve the limitations of UDA, they are designed for classification, and they are not directly applicable to re-id, which is an open set task where the set of personal identities differs between the source and target domains.
For re-id, BNTA~\cite{han2022generalizable} introduces a self-supervised training that leverages the body structure of human (\eg, head and foot are in upper and lower regions in images) in feature representation.
However, BNTA requires special modifications to the model architecture and source-pretraining procedure for the self-supervised training, which limits reuse of off-the-shelf models.

Inspired by uncertainty minimization~\cite{Wang2021,zhou2021bayesian,niu2022efficient}, we propose a novel FTTA method for re-id, \emph{test-time similarity modification for person re-identification ({\proposedmethod})}.
{\proposedmethod} is the first FTTA method for re-id that can reuse pre-trained re-id models trained with arbitrary methods and perform the adaptation in an online manner.
\section{Problem Setting}\label{sec:problem_setting}
\subsection{Person Re-identification}
In re-id, a feature extractor $f_\theta: \mathcal{X} \to \mathbb{R}^d$ is trained to make the images of the same person form a cluster in the feature space, where $\mathcal{X}$ is the input space.
The goal of re-id at test time is, given a query person image $\mathbf{x}^\text{q} \in \mathcal{X}$, to retrieve the person images from gallery images $\{ \mathbf{x}^\text{g}_i \}_{i=1}^n \in \mathcal{X}^n$ that have the same identity as the query person.
Since the person identities of the test set differ from the training ones in re-id, the retrieval from the galleries is based on the similarity or distance in the feature space of $f_\theta$.

\subsection{Fully Test-time Adaptation in Person Re-identification}
Given a source-pretrained feature extractor $f_{\theta_0}$, we aim to update it to maintain the accuracy of re-id in the target domain where the distribution differs from that of the source domain and further changes over time.
That is, given a batch of incoming query person images $\mathcal{B}=\{ \mathbf{x}^\text{q}_1 ,\ldots, \mathbf{x}^\text{q}_{|\mathcal{B}|} \} \in \mathcal{X}^{|\mathcal{B}|}$ from the target domain, we seek to find the images of persons identical to the query ones from the gallery  $\mathcal{G}=\{ \mathbf{z}^\text{g}_i \}_{i=1}^n$ through their similarity in the feature space. Here, $\mathbf{z}^\text{g}_i = f_{\theta_0} (\mathbf{x}^\text{g}_i)$ is a pre-extracted gallery feature, and $\mathbf{x}^\text{g}_i$ is a gallery person image.
Since extracting features from the gallery person images for every comparison to the query features is computationally inefficient, we assume that the gallery features are pre-extracted and fixed.
This assumption is reasonable in re-id, whose task is to retrieve images from the gallery stored in a database.

\section{Test-time Similarity Modification for Person Re-identification}\label{sec:test-time_similarity_tuning}
\inputfig[*]{0.7}{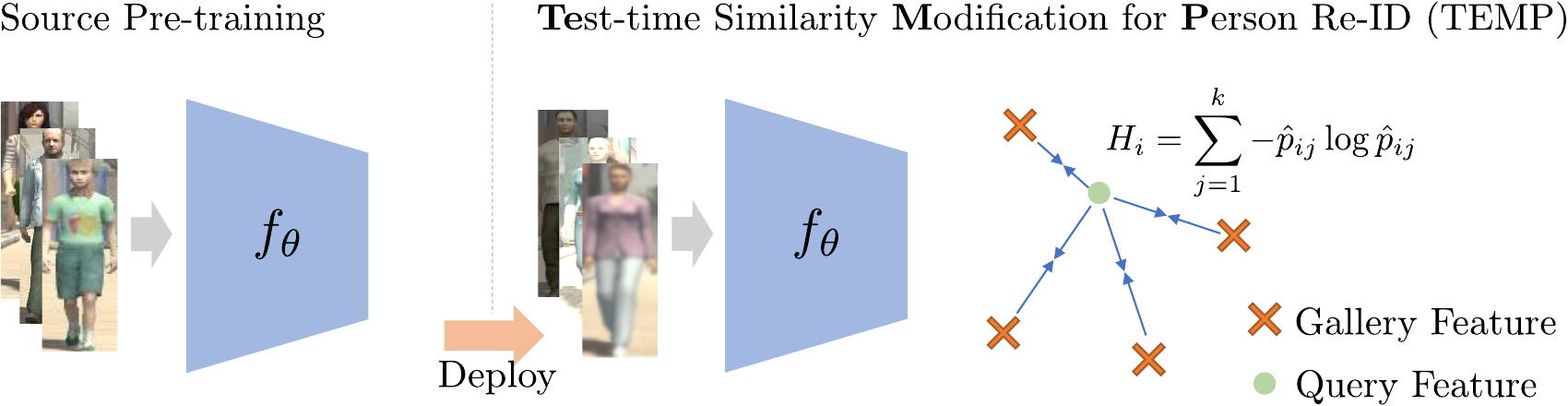}{Overview of {\proposedmethod}.
{\proposedmethod} alleviates the drop in performance of a re-id model caused by a distribution shift by using re-id entropy to modify the similarities between the query and gallery features.
The person images in the figure are sampled from PersonX~\cite{Sun_2019_CVPR}.}{fig:test_overview}{Overview of {\proposedmethod}.}

\cref{fig:test_overview} is an overview of test-time similarity modification for person re-identification ({\proposedmethod}).
{\proposedmethod} aims at minimizing the uncertainty of the prediction of re-id during testing.
The previous TTA methods that take this approach for classification~\cite{Wang2021,zhou2021bayesian,niu2022efficient} improve accuracy on the target domain.
In this section, we briefly describe the uncertainty minimization approach for the classification models and then introduce {\proposedmethod}, which is for re-id models.

In classification where the class set is shared between the source and target domains, the uncertainty of the prediction $\hat{p}(y|x)$ can be measured via entropy:
\begin{equation}
H = \sum_{y=1}^C -\hat{p}(y|x) \log \hat{p}(y|x),\label{eq:classification_entropy}
\end{equation}
where $x$ is an input and $y\in \{ 1,\ldots,C \}$ is the class.
Wang \etal~\cite{Wang2021} found that entropy strongly correlates with accuracy on the target domain, \ie, the entropy increases when the accuracy deteriorates due to a distribution shift.
Thus, to improve accuracy, they minimize entropy on the target data at test time.

On the other hand, unlike classification, the outputs of $f_\theta$ are feature vectors, not probabilities, since re-id does not have classes that are shared by the source and target domains.
Instead, we regard each gallery feature as corresponding to a class.
Although two or more gallery features may correspond to the same person identity, we treat each gallery feature as distinct for simplicity.
However, unlike the classification case, the set of ``classes'' varies depending on the galleries $\mathcal{G}$.
Thus, based on the feature similarity, we compute the probability that query person $\mathbf{x}^\text{q}_i$ from the target domain is the same as the $j$-th gallery image $\mathbf{x}^\text{g}_j ~ (j\in \{ 1,\ldots, n \})$.

To compute the probability, first, we compute cosine similarity between the query feature $\mathbf{z}^\text{q}_i = f_\theta (\mathbf{x}^\text{q}_i)$ and each gallery feature $\mathcal{G}=\{ \mathbf{z}^\text{g}_j \}_{j=1}^n$:
\begin{equation}
    s_{ij} = \cos_\text{s} \left(\mathbf{z}^\text{q}_i, \mathbf{z}^\text{g}_j \right) = \frac{{\mathbf{z}^\text{q}_i} \cdot \mathbf{z}^\text{g}_j}{\| \mathbf{z}^\text{q}_i \|_2 \| \mathbf{z}^\text{g}_j \|_2}. \label{eq:cos_similarity}
\end{equation}

Second, we select the top-$k$ similar gallery features in terms of $s_{ij}$ denoted by $\{ \mathbf{z}^\text{g}_{a_{ij'}} \}_{j'=1}^k$, where $a_{ij'}\in\{ 1,\ldots,n \}$ is the gallery index selected for $\mathbf{z}_i^\text{q}$.
This is because the number of galleries $n$ may be too large (\eg, tens of thousands) to compute the probability described below.

Then, within the top-$k$ items, we regard the similarity as the logit in classification and compute the probability with the softmax function:
\begin{equation}
    \hat{p}_{ij} = \frac{\exp ( s_{i,a_{ij}} )}{\sum_{j'=1}^k \exp ( s_{i,a_{ij'}} )}.
    \label{eq:proposed_softmax}
\end{equation}

This enables us to compute the entropy analogously to the classification case in \cref{eq:classification_entropy}:
\begin{equation}
H_i = \sum_{j=1}^k -\hat{p}_{ij} \log \hat{p}_{ij}.
\label{eq:re-id_entropy}
\end{equation}

We refer to it as \emph{re-id entropy}.
We minimize re-id entropy in order to adapt the feature extractor $f_\theta$ to the target domain.
We experimentally found that re-id entropy also tends to correlate with performance in the case of a distribution shift (see \cref{sssec:exp_entropy}).
By minimizing the re-id entropy, the similarities between the query and gallery features are ``modified'' to make the query features move toward the nearest gallery ones.

Additionally, inspired by a fine-tuning method L2-SP~\cite{pmlr-v80-li18a}, we add a L2 regularization between the current and source parameters $\theta$ and $\theta_0$ to avoid forgetting the knowledge learned from the source domain.

Finally, our objective is:
\begin{equation}
    \min_\theta \frac{1}{|\mathcal{B}|}\sum_{i=1}^{|\mathcal{B}|} H_i + \lambda \| \theta - \theta_0 \|_2^2,
    \label{eq:proposed_objective}
\end{equation}
where $\lambda > 0$ is a hyperparameter to determine the effect of the second term for anti-forgetting.

Following Tent~\cite{Wang2021}, we optimize only the affine parameters $\gamma$ and $\beta$ in the BN layers instead of all the parameters to avoid forgetting.
The detailed procedure of {\proposedmethod} is given in \cref{alg:temp}.

\renewcommand{\algorithmicrequire}{\textbf{Input:}}
\renewcommand{\algorithmicensure}{\textbf{Output:}}

\begin{algorithm}
\caption{Test-time similarity modification for person re-identification ({\proposedmethod}).}
\label{alg:temp}
{
\begin{algorithmic}
\REQUIRE{Source-pretrained feature extractor $f_{\theta_0}$, gallery features $\mathcal{G}=\{ \mathbf{z}^\text{g}_i \}_{i=1}^n$, batches of query images $\{ \mathcal{B} \}$, number of gallery features to be used $k$, regularization weight $\lambda$, and learning rate $\eta$}
\ENSURE{Adapted feature extractor $f_{\theta}$ and re-id results}
\FOR{$\mathcal{B}$ in query batches}
\STATE{Extract query features $\{ \mathbf{z}^\text{q}_i \}_{i=1}^{|\mathcal{B}|}$.}
\STATE{Compute similarity $s_{ij}$ between $\mathbf{z}^\text{q}_i$ and $\mathbf{z}^\text{g}_j$ as \cref{eq:cos_similarity}.}
\STATE{Perform re-id based on $s_{ij}$ for current query batch $\mathcal{B}$.}
\STATE{Select top-$k$ similar gallery features $\{ \mathbf{z}_{a_{i1}}^\text{g}, \ldots, \mathbf{z}_{a_{ik}}^\text{g} \}$ for each $\mathbf{z}_i^\text{q}$.}
\STATE{Compute probability as \cref{eq:proposed_softmax}.}
\STATE{Compute re-id entropy as \cref{eq:re-id_entropy}.}
\STATE{Update $\theta$ toward \cref{eq:proposed_objective}:\\$\theta \leftarrow \theta - \eta \nabla_\theta \left[ (1/|\mathcal{B}|)\sum_{i=1}^{|\mathcal{B}|} H_i +\lambda \| \theta-\theta_0 \|_2^2 \right]$.}
\ENDFOR
\end{algorithmic}
}
\end{algorithm}
\section{Experiment}\label{sec:experiment}
We carried out experiments with {\proposedmethod} and the baselines on re-id tasks where the target domain distribution changes over time.
We confirmed that distribution shifts degrade the performance of re-id and increase re-id entropy defined in \cref{eq:re-id_entropy} (\cref{sssec:exp_entropy}) and that {\proposedmethod} alleviates drops in accuracy caused by distribution shifts (\cref{sssec:result_domain-change,sssec:result_condition-change,sssec:sensitivity_analysis}).
Moreover, we observed that {\proposedmethod} reduces the distribution gap between the source and target domains in the feature space (\cref{sssec:feature_visualization}).

\subsection{Experimental Setting}\label{ssec:exp_setting}
\subsubsection{Dataset}\label{sssec:exp_dataset}
We selected three datasets that have been widely used in person re-id studies: Market-1501~\cite{Zheng_2015_ICCV}, MSMT17~\cite{Wei_2018_CVPR}, and PersonX~\cite{Sun_2019_CVPR}.

Market-1501 contains 32,668 pedestrian images taken from six cameras, including 1,501 identities.
MSMT17 contains 126,441 images taken from 15 cameras, including 4,101 identities.
PersonX is a dataset of synthetic person images different from the aforementioned ones.
It contains 273,456 synthetic person images of 1,266 identities rendered from 36 viewpoints.
These datasets are divided into the training and test sets, and the test sets are further split into query and gallery sets.

We considered two scenarios involving temporal distribution shifts: (i) location change and (ii) image corruption.
We show that {\proposedmethod} can adapt to both types of temporal shifts in an online manner.
\\
\textbf{Location change} is a scenario where the domains of both the query and gallery images change over time.
This situation can occur when a trained re-id model or system is carried to other locations, \eg, in-vehicle cameras.
In this scenario, the distributions of queries and galleries change since different locations generate different person images. 
We simulated location changes by altering the above datasets one after another. \\
\textbf{Image corruption} is a scenario where the distribution of the query images changes over time while the gallery distribution is fixed.
This situation can occur when images are collected from fixed-point cameras in offices or on city streets, for example.
The distribution of the query images changes with the environment over time, \eg, the light or image quality may vary.
We added corruptions to the query images and changing their strength gradually.
We tested three types of corruption: brightness (\eg, daytime and nighttime), Gaussian blur (\eg, out of focus), and pixelate (\eg, image resolution degradation).

\subsubsection{Pre-training of the Re-id Model}\label{sssec:source_pretraining}
We trained ResNet-50~\cite{resnet} for the source-pretrained models on the training sets of the three datasets.
As usually done in re-id, we minimized the cross-entropy and softmax-triplet~\cite{Qian_2019_ICCV} loss functions.
For optimization, we used Adam~\cite{adam} and set the learning rate to 0.00035, batch size to 16, weight decay to 0.0005, and the number of epochs to 120.
We implemented the pre-training and TTA algorithms on OpenUnReID~\cite{openunreid}.
For BNTA~\cite{han2022generalizable}, we added the self-supervised training branch to the model following the paper.
We performed the training and the below TTA with a single NVIDIA A100 GPU (40 GiB).

\begin{figure*}[tb]
\centering
\begin{tabular}{ccc}
\includegraphics[alt={Top-1 CMC accuracy, mean re-id entropy, and corruption strength (Brightness)},width=0.3\linewidth]{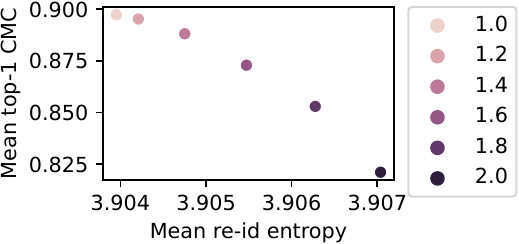} &
\includegraphics[alt={Top-1 CMC accuracy, mean re-id entropy, and corruption strength (Gaussian blur)},width=0.3\linewidth]{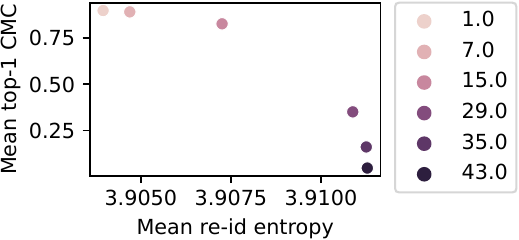} &
\includegraphics[alt={Top-1 CMC accuracy, mean re-id entropy, and corruption strength (Pixelate)},width=0.3\linewidth]{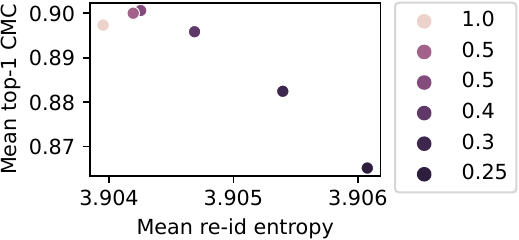} \\
{\footnotesize Brightness} & {\footnotesize Gaussian blur} & {\footnotesize Pixelate}
\end{tabular}
\caption{Top-1 CMC accuracy, mean re-id entropy, and corruption strength on Market-1501.
As the corruption strength (represented by the color of the circle, darker is stronger) increases, the re-id entropy also increases, and the top-1 CMC accuracy deteriorates.}
\label{fig:corruption_entropy}
\end{figure*}

\subsubsection{Test-time Adaptation}\label{sssec:test-time_adaptation}
For {\proposedmethod}, we optimized the loss with \cref{eq:proposed_objective}.
We used the same optimizer and hyperparameters as in the pre-training except for setting the weight decay to 0.
We set $k$ to 50, determining the number of gallery features to be used in \cref{eq:proposed_softmax}, and $\lambda$ in \cref{eq:proposed_objective} to 0.0001.
We did not update the BN statistics (feature mean and standard deviation) during TTA because we found that updating them degrades re-id performance.
As we consider the online setting, for each mini-batch of the query images, we made predictions, updated the model one step, and then discarded the mini-batch.

In the location change scenario, we ran TTA while altering the test splits of the three datasets one after another (\eg, PersonX $\rightarrow$ MSMT17 $\rightarrow$ Market-1501 when the source model is pre-trained on Market-1501).
When the datasets (phases) switched, we extracted the gallery features $\{ \mathbf{z}_i^\text{g} \}_i$ from the current dataset and then fed the query images to the model.

In the image corruption scenario, we used the same dataset as in the pre-training but added corruptions to the query images.
Before TTA, we extracted the gallery features from the images without corruption and fixed them during TTA.
We gradually increased the strength of the corruptions added to the query images.
We ran TTA for one epoch for each strength of corruption and then switched to the next strength.
We tested TTA with brightness, Gaussian blur, and pixelate.
We gradually strengthened the corruptions in five steps for Gaussian blur and pixelate.
For brightness, we made the images darker and brighter in eight steps.

We compared {\proposedmethod} with No-adapt, BN-adapt~\cite{Benz_2021_WACV}, SourceTent, and BNTA~\cite{han2022generalizable}.
\\
\textbf{No-adapt} means just testing the source-pretrained model without any model updates and modifications.
\\
\textbf{BN-adapt}~\cite{Benz_2021_WACV} keeps updating the statistics of the BN layers during testing.
BN-adapt estimates the mean and variance of the features from each mini-batch.
\\
\textbf{SourceTent} is a naive way to apply Tent~\cite{Wang2021}, which is a representative TTA method for classification, to re-id.
SourceTent employs the last fully-connected classification layer used during the source-pretraining.
Although the set of person identity in test time is different from that in the source-pretraining.
SourceTent computes logits with the classification layer and minimizes the prediction entropy in \cref{eq:classification_entropy}.
Note that SourceTent can be applied only when the source-pretraining is done in a supervised manner, while unsupervised training is also used in re-id~\cite{lin2021unsupervised,re-id_uda_survey}.
\\
\textbf{BNTA}~\cite{han2022generalizable} is a TTA method developed for re-id.
It employs a self-supervised training for leveraging the body structure of human (\eg, head and foot are in upper and lower regions in images) in feature representation.
Although the gallery images are used for TTA in the original paper~\cite{han2022generalizable}, we used the query images since our problem setting assumes that the gallery features are fixed and mini-batches of the query images come sequentially. 

We used top-1 cumulative matching characteristics (CMC) accuracy for evaluation.
We ran each TTA with different random seeds three times and report the average and standard deviations.

\subsection{Results}\label{ssec:exp_result}

\subsubsection{Distribution Shift and Re-id Entropy}\label{sssec:exp_entropy}
To verify that re-id entropy measures the degree of distribution shift in re-id, we computed re-id entropy defined in \cref{eq:re-id_entropy} with the source model $f_{\theta_0}$.
We vary the strengths of the corruption of the query images.
\cref{fig:corruption_entropy} shows the relationship between the mean of re-id entropy, top-1 CMC accuracy, and corruption strength.
As the corruption becomes stronger, re-id entropy increases, and the performance of re-id deteriorates.
Thus, re-id entropy is a good proxy of the performance of re-id and we can expect that reducing re-id entropy will improve the performance of re-id.

\subsubsection{Results for Location Change}\label{sssec:result_domain-change}

\def\plotwidth{0.41\linewidth}

\begin{figure*}[tb]
\centering
\begin{tabular}{cc}
\includegraphics[alt={Top-1 CMC accuracy vs. iteration (Location change, starts from PersonX).},width=\plotwidth]{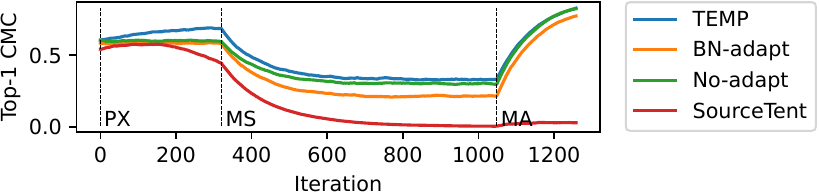}
&
\includegraphics[alt={Top-1 CMC accuracy vs. iteration (Corruption, brightness).},width=\plotwidth]{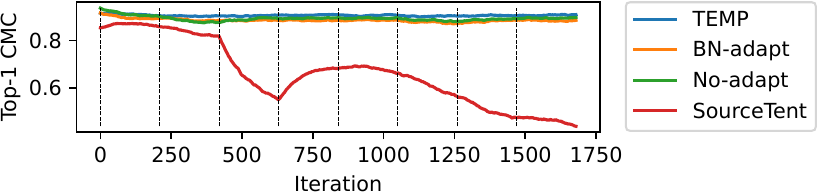}
\\
\includegraphics[alt={Top-1 CMC accuracy vs. iteration (Location change, starts from Market-1501).},width=\plotwidth]{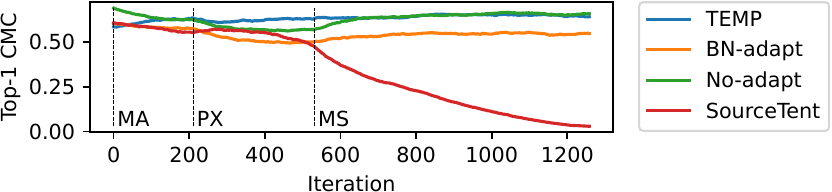}
&
\includegraphics[alt={Top-1 CMC accuracy vs. iteration (Corruption, Gaussian blur).},width=\plotwidth]{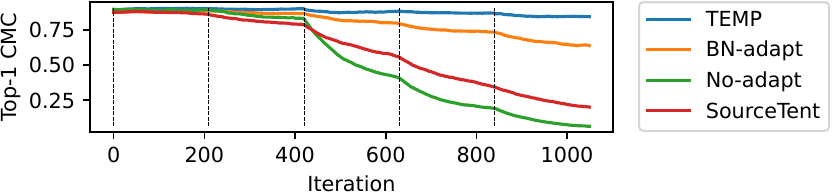}
\\
\includegraphics[alt={Top-1 CMC accuracy vs. iteration (Location change, starts from MSMT17).},width=\plotwidth]{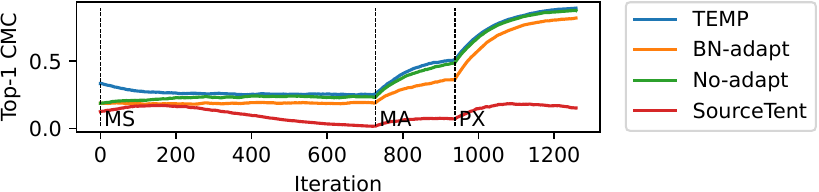}
&
\includegraphics[alt={Top-1 CMC accuracy vs. iteration (Corruption, pixelate).},width=\plotwidth]{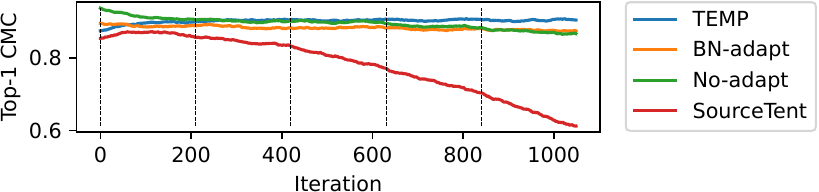}
\\
{\footnotesize (a) Location change} & {\footnotesize (b) Corruption}
\end{tabular}
\caption{Top-1 CMC accuracy {\vs} iteration.
The exponential moving average of the CMCs computed for each mini-batch is plotted.
\textbf{(a)}: The location change scenario. The source datasets used to plot the top, middle, and bottom rows are Market-1501, MSMT17, and PersonX, respectively.
The vertical dashed lines represent the iterations when the dataset switches.
MA, MS, and PX in the figure represent Market-1501, MSMT17, and PersonX.
\textbf{(b)}: The corruption scenario.
The source dataset is Market-1501.
The top, middle, and bottom rows correspond to  brightness, Gaussian blur, and pixelate corruptions.
The vertical dashed lines represent the iterations when the strengths of the corruptions change.
BNTA~\cite{han2022generalizable} is not plotted since it immediately decreased the accuracy to near to zero.
}
\label{fig:cmc_transitions}
\end{figure*}

\begin{table}[tb]
\centering
\caption{Average top-1 CMC accuracy in the location change scenario.}
\label{tab:domain_online_cmc}
{\footnotesize
\setlength{\tabcolsep}{3pt}
\begin{tabular}{lllll}\toprule
~ & ~ & \multicolumn{3}{c}{Phase} \\
Source & Method & 1 & 2 & 3  \\ \midrule
\multirow{6}{*}{Market-1501} & ~ & PersonX & MSMT17 & Market-1501 \\
~ & No-adapt & $59.74_{\pm 0.00}$ & $29.91_{\pm 0.00}$ & $\mathbf{89.76_{\pm 0.00}}$ \\
~ & BN-adapt~\cite{Benz_2021_WACV} & $58.33_{\pm 0.24}$ & $21.12_{\pm 0.30}$ & $84.85_{\pm 0.26}$ \\
~ & SourceTent & $51.31_{\pm 0.55}$ & $1.97_{\pm 0.27}$ & $3.58_{\pm 0.82}$ \\
~ & BNTA~\cite{han2022generalizable} & $1.54_{\pm 0.18}$ & $0.13_{\pm 0.07}$ & $0.75_{\pm 0.47}$ \\
~ & TEMP & $\mathbf{68.30_{\pm 0.35}}$ & $\mathbf{32.89_{\pm 0.30}}$ & $89.33_{\pm 0.16}$ \\ \midrule
\multirow{6}{*}{MSMT17} & ~ & Market-1501 & PersonX & MSMT17 \\
~ & No-adapt & $61.22_{\pm 0.00}$ & $55.92_{\pm 0.00}$ & $\mathbf{65.42_{\pm 0.00}}$ \\
~ & BN-adapt~\cite{Benz_2021_WACV} & $56.98_{\pm 0.31}$ & $49.10_{\pm 0.20}$ & $54.55_{\pm 0.37}$ \\
~ & SourceTent & $55.23_{\pm 0.53}$ & $52.04_{\pm 0.91}$ & $12.67_{\pm 1.17}$ \\
~ & BNTA~\cite{han2022generalizable} & $3.29_{\pm 0.30}$ & $5.96_{\pm 1.76}$ & $0.81_{\pm 0.49}$ \\
~ & TEMP & $\mathbf{63.62_{\pm 0.24}}$ & $\mathbf{61.93_{\pm 0.25}}$ & $64.47_{\pm 0.17}$ \\ \midrule
\multirow{6}{*}{PersonX} & ~ & MSMT17 & Market-1501 & PersonX \\
~ & No-adapt & $23.35_{\pm 0.00}$ & $51.73_{\pm 0.00}$ & $89.41_{\pm 0.00}$ \\
~ & BN-adapt~\cite{Benz_2021_WACV} & $18.75_{\pm 0.05}$ & $38.56_{\pm 0.14}$ & $83.98_{\pm 0.43}$ \\
~ & SourceTent & $8.71_{\pm 0.12}$ & $8.76_{\pm 1.12}$ & $18.47_{\pm 0.15}$ \\
~ & BNTA~\cite{han2022generalizable} & $0.40_{\pm 0.14}$ & $3.53_{\pm 2.50}$ & $7.63_{\pm 4.72}$ \\
~ & TEMP & $\mathbf{25.34_{\pm 0.19}}$ & $\mathbf{54.55_{\pm 0.26}}$ & $\mathbf{90.69_{\pm 0.28}}$ \\ \bottomrule
\end{tabular}
}
\end{table}

We sequentially evaluated top-1 CMC accuracy on each mini-batch and averaged the accuracies for each phase during testing.
\cref{tab:domain_online_cmc} and \cref{fig:cmc_transitions} (a) show the average top-1 CMC accuracies in each phase and the transition of CMC accuracy.
Each phase corresponds to a dataset.
{\proposedmethod} alleviated the performance drop and had the best CMC accuracy in most cases.
In contrast, BN-adapt~\cite{Benz_2021_WACV}, a simple and effective TTA method for classification, showed a significant drop in performance.
This is because distribution shifts due to location changes are more complex than image corruptions, and the batch size (16) is smaller than the ones ordinarily used in classification (\eg, 64, 128).
In the online setting where queries come sequentially, increasing the batch size is unsuitable because waiting for the data to arrive can cause a delay.
Thus, BN-adapt is not effective in this setting because it is hard to estimate BN statistics accurately with a small batch~\cite{zhao2023delta}.
SourceTent also significantly drops the accuracy.
This is because the last fully connected layer for classification in source-pretraining is used for computing entropy.
As re-id is an open-set task, the set of person identities in test time differs that used in the source-pretraining.
Thus, minimizing the entropy within the source identity set for test data resulted in the accuracy drop.
BNTA~\cite{han2022generalizable} also significantly drops the accuracy in our setting.
This is because BNTA is not designed for the online TTA like {\proposedmethod}.
For TTA, BNTA requires sufficient steps of self-supervised training with test data, while the model has to adapt in an online manner in our setting.

\subsubsection{Results for Image Corruptions}\label{sssec:result_condition-change}
\cref{fig:cmc_transitions} (b) shows the top-1 CMC accuracy tested on the three corruptions of Market-1501.
{\proposedmethod} maintained its level of the performance as the strength of the corruption increased even when the baselines showed significant drops in performance.

\subsubsection{Effect of the Hyperparameter}\label{sssec:sensitivity_analysis}
\inputfig{1}{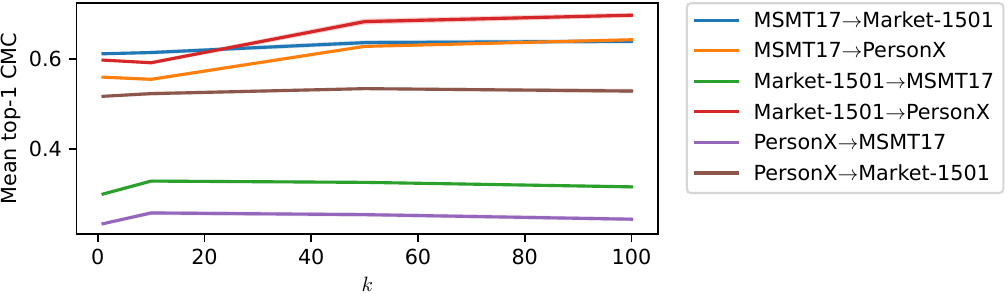}{Sensitivity analysis of $k$.}{fig:top-k_vs_mAP}{Sensitivity analysis of k.}

{\proposedmethod} has a hyperparameter $k$ that determines the number of gallery features to be used in \cref{eq:proposed_softmax}.
To examine the effect of $k$, we varied $k$ within $\{ 1,10,50,100 \}$ and performed TTA in the location change scenario.
\cref{fig:top-k_vs_mAP} shows the top-1 CMC accuracies after running {\proposedmethod} on the six pairs of the source and target datasets.
The top-1 CMC accuracy tended to grow with a larger $k$ but decreased when $k$ exceeded 50.
Hence, we consider that setting $k$ to about 50 strikes the best balance between performance and computational cost, as the cost of computing the probability in \cref{eq:proposed_softmax} is proportional to $k$.

\subsubsection{Effect of the Similarity Modification}\label{sssec:feature_visualization}

\begin{figure}[tb]
\centering
\begin{tabular}{cc}
\includegraphics[alt={Visualization of the query and gallery features in the corruption scenario (No-adapt).},width=0.4\linewidth]{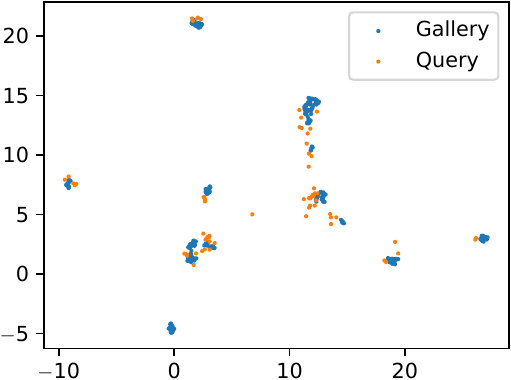} & 
\includegraphics[alt={Visualization of the query and gallery features in the corruption scenario ({\proposedmethod}).},width=0.4\linewidth]{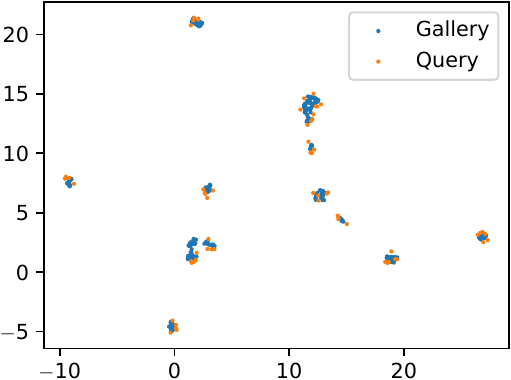} \\
 {\footnotesize No-adapt} & {\footnotesize \proposedmethod}
\end{tabular}
\caption{UMAP~\cite{mcinnes2018umap-software} visualization of the query and gallery features in the corruption scenario.
The gallery images are not corrupted, and the query ones are Gaussian-blurred.
{\proposedmethod} aligns the distribution of the gallery (blue dots) and query (orange dots) features.}
\label{fig:feature_umap}
\end{figure}

To confirm the effect of TEMP from the perspective of the feature space, we visualized the query and gallery features in the corruption scenario.
\cref{fig:feature_umap} shows the UMAP~\cite{mcinnes2018umap-software} visualization of the gallery and query features.
We used images of 30 randomly sampled person identities.
Since the gallery features are fixed while the query feature distribution changes, we extracted the features from the gallery images without corruption and trained a UMAP mapper on the gallery features.
Then, we embedded the gallery and query features with the mapper.
As shown in \cref{fig:feature_umap}, {\proposedmethod} aligned the clusters of the query and gallery features.
This implies that {\proposedmethod} pulls the query features to the corresponding gallery ones by similarity modification, which helps re-id during testing.

\subsection{Effect of the Batch Size} \label{ssec:batch_size_effect}
\begin{table*}[tb]
\centering
\caption{Average top-1 CMC accuracy {\vs} batch size in the location change scenario.
Each CMC accuracy is the mean over the three phases.
Methods other than {\proposedmethod} and No-adapt cannot run when the batch size is one since the BN statistics cannot be updated.}
\label{tab:cmc_location_batch-size}
{\footnotesize
\begin{tabular}{lllllllll}\toprule
~ & ~ & \multicolumn{7}{c}{Batch size} \\
Source & Method & 64 & 32 & 16 & 8 & 2 & 1 & 1 (lr=3.5e-5) \\ \midrule
\multirow{5}{*}{Market-1501} & No-adapt & $47.44_{\pm 0.00}$ & $47.51_{\pm 0.00}$ & $47.50_{\pm 0.00}$ & $47.51_{\pm 0.00}$ & $47.51_{\pm 0.00}$ & $47.51_{\pm 0.00}$ & - \\
~ & BN-adapt~\cite{Benz_2021_WACV} & $42.18_{\pm 0.04}$ & $41.92_{\pm 0.08}$ & $41.24_{\pm 0.24}$ & $39.58_{\pm 0.30}$ & $27.88_{\pm 0.63}$ & - & - \\
~ & SourceTent & $24.59_{\pm 0.09}$ & $19.16_{\pm 0.27}$ & $14.82_{\pm 0.28}$ & $10.16_{\pm 0.16}$ & $4.00_{\pm 0.08}$ & - & - \\
~ & BNTA~\cite{han2022generalizable} & $1.82_{\pm 0.16}$ & $1.12_{\pm 0.23}$ & $0.60_{\pm 0.07}$ & $0.42_{\pm 0.08}$ & - & - & - \\
~ & TEMP & $\mathbf{51.00_{\pm 0.05}}$ & $\mathbf{51.23_{\pm 0.09}}$ & $\mathbf{51.33_{\pm 0.13}}$ & $\mathbf{51.25_{\pm 0.19}}$ & $\mathbf{50.96_{\pm 0.27}}$ & ${41.11_{\pm 13.31}}$ & $\mathbf{51.03_{\pm 0.14}}$ \\ \midrule
\multirow{5}{*}{MSMT17} & No-adapt & $62.33_{\pm 0.00}$ & $62.32_{\pm 0.00}$ & $62.30_{\pm 0.00}$ & $62.29_{\pm 0.00}$ & $62.30_{\pm 0.00}$ & $62.30_{\pm 0.00}$ & - \\
~ & BN-adapt~\cite{Benz_2021_WACV} & $55.03_{\pm 0.10}$ & $54.64_{\pm 0.11}$ & $53.57_{\pm 0.24}$ & $51.64_{\pm 0.35}$ & $32.78_{\pm 0.84}$ & - & - \\
~ & SourceTent & $51.56_{\pm 0.18}$ & $46.29_{\pm 0.37}$ & $29.81_{\pm 0.97}$ & $18.16_{\pm 0.86}$ & $6.90_{\pm 0.64}$ & - & - \\
~ & BNTA~\cite{han2022generalizable} & $3.29_{\pm 0.37}$ & $3.26_{\pm 0.52}$ & $2.54_{\pm 0.67}$ & $0.84_{\pm 0.04}$ & - & - & - \\
~ & TEMP & $\mathbf{63.22_{\pm 0.09}}$ & $\mathbf{63.41_{\pm 0.11}}$ & $\mathbf{63.68_{\pm 0.04}}$ & $\mathbf{63.65_{\pm 0.22}}$ & $\mathbf{63.34_{\pm 0.04}}$ & $\mathbf{62.75_{\pm 0.34}}$ & $\mathbf{63.61_{\pm 0.04}}$ \\ \midrule
\multirow{5}{*}{PersonX} & No-adapt & $44.87_{\pm 0.00}$ & $44.89_{\pm 0.00}$ & $44.93_{\pm 0.00}$ & $44.92_{\pm 0.00}$ & $44.92_{\pm 0.00}$ & $44.93_{\pm 0.00}$ & - \\
~ & BN-adapt~\cite{Benz_2021_WACV} & $39.56_{\pm 0.07}$ & $39.28_{\pm 0.03}$ & $38.68_{\pm 0.15}$ & $37.47_{\pm 0.03}$ & $26.50_{\pm 0.28}$ & - & - \\
~ & SourceTent & $31.38_{\pm 0.05}$ & $18.01_{\pm 0.04}$ & $11.21_{\pm 0.29}$ & $6.39_{\pm 0.09}$ & $1.41_{\pm 0.05}$ & - & - \\
~ & BNTA~\cite{han2022generalizable} & $1.77_{\pm 0.46}$ & $1.35_{\pm 0.34}$ & $2.77_{\pm 1.69}$ & $0.27_{\pm 0.06}$ & - & - & - \\
~ & TEMP & $\mathbf{46.77_{\pm 0.02}}$ & $\mathbf{46.80_{\pm 0.09}}$ & $\mathbf{46.88_{\pm 0.05}}$ & $\mathbf{46.79_{\pm 0.14}}$ & $\mathbf{46.30_{\pm 0.18}}$ & ${7.12_{\pm 2.65}}$ & $\mathbf{46.78_{\pm 0.07}}$ \\ \bottomrule
\end{tabular}
}
\end{table*}

In re-id systems, person images often come sequentially.
Thus, models may have to predict person images one after another rather than waiting to accumulate data for making a mini-batch.
We ran TTA with varying the batch size within $\{ 64,32,16,8,2,1 \}$ in the location change scenario.

\cref{tab:cmc_location_batch-size} shows the result.
BN-adapt~\cite{Benz_2021_WACV} has low accuracy when the batch size is small because smaller batch size results in an inaccurate estimation of BN statistics.
Even when the batch size is increased, BN-adapt underperforms No-adapt.
This is because re-id has many galleries such as tens of thousands in the experiments (corresponding to each classes in classification), which is much larger than the batch size.
The other baselines also degrade the accuracy especially when the batch size is small since they depend on the batch statistics.
As each mini-batch contains different set of person identity (class) and its statistics are unstable, estimating BN statistics that fit all the galleries from the batches is difficult.
On the other hand, {\proposedmethod} consistently outperformed the baselines regardless of the batch size since it does not update the BN statistics.
When the batch size is one, {\proposedmethod} drops the accuracy.
This is because the model overfits since the total number of iterations increases as the batch size decreases.
In such case, reducing the learning rate to 3.5e-5 improved the accuracy.

\subsection{Inference Time} \label{ssec:computational_cost}
\begin{table}[tb]
\centering
\caption{Inference time (msec/image) during TTA.}
\label{tab:inference_time}
{
\begin{tabular}{llllll} \toprule
~ & \multicolumn{5}{c}{Batch size} \\
Method & 32 & 16 & 8 & 2 & 1 \\ \midrule
No-adapt & $22.58$ & $26.97$ & $35.43$ & $87.48$ & $159.1$ \\
BN-adapt~\cite{Benz_2021_WACV} & $22.44$ & $26.81$ & $35.58$ & $89.22$ & - \\
SourceTent & $25.65$ & $28.52$ & $42.75$ & $106.5$ & - \\
BNTA~\cite{han2022generalizable} & $24.89$ & $29.34$ & $38.03$ & - & - \\
TEMP & $26.33$ & $31.64$ & $43.19$ & $111.2$ & $204.8$ \\ \bottomrule
\end{tabular}
}

\end{table}

Inference time is an important factor for re-id as real-time inference is often required.
We compared the inference time (msec/image) during TTA.
Here we performed each TTA method on MSMT17.
\cref{tab:inference_time} shows the result.
In all methods, the inference time increases as the batch size decreases because the overhead (\eg, communicating with GPUs) increases.
Compared to the baselines, {\proposedmethod} slightly increases the inference time.
But especially in larger batch size, the difference of the inference time gets smaller.
The inference cost of {\proposedmethod} can be decreased further by simple modifications, \eg, updating the parameters only once every a certain number of iterations, or selecting data to be used in model update like EATA~\cite{niu2022efficient}.

\section{Conclusion}\label{sec:conclusion}
We proposed test-time similarity modification for person re-identification ({\proposedmethod}), a simple yet effective Fully TTA method for re-id.
{\proposedmethod} can be applied in an online manner and can alleviate drop in the performance of re-id caused by temporal distribution shifts.
We examined two scenarios in the experiments: location changes and image corruption.
{\proposedmethod} outperformed the baselines in both scenarios, and we found that the query and gallery feature distributions were aligned via similarity modification.
One limitation of {\proposedmethod} is that it sometimes slightly underperforms when the distribution returns to the source one.
Since {\proposedmethod} instantly adapts to the current distribution, it has to adapt to the source domain again.
However, {\proposedmethod} can be easily extended to incorporate other techniques, \eg, improved anti-forgetting regularization of BACS~\cite{zhou2021bayesian} or EATA~\cite{niu2022efficient}.
Improving {\proposedmethod} by incorporating such techniques is one direction of our future work.

\clearpage
\bibliographystyle{IEEEtran}
\bibliography{main}

\begin{thebibliography}{10}
\providecommand{\url}[1]{#1}
\csname url@samestyle\endcsname
\providecommand{\newblock}{\relax}
\providecommand{\bibinfo}[2]{#2}
\providecommand{\BIBentrySTDinterwordspacing}{\spaceskip=0pt\relax}
\providecommand{\BIBentryALTinterwordstretchfactor}{4}
\providecommand{\BIBentryALTinterwordspacing}{\spaceskip=\fontdimen2\font plus
\BIBentryALTinterwordstretchfactor\fontdimen3\font minus \fontdimen4\font\relax}
\providecommand{\BIBforeignlanguage}[2]{{%
\expandafter\ifx\csname l@#1\endcsname\relax
\typeout{** WARNING: IEEEtran.bst: No hyphenation pattern has been}%
\typeout{** loaded for the language `#1'. Using the pattern for}%
\typeout{** the default language instead.}%
\else
\language=\csname l@#1\endcsname
\fi
#2}}
\providecommand{\BIBdecl}{\relax}
\BIBdecl

\bibitem{alexnet}
A.~Krizhevsky, I.~Sutskever, and G.~E. Hinton, ``Imagenet classification with deep convolutional neural networks,'' in \emph{Advances in neural information processing systems}, 2012.

\bibitem{Simonyan2015VeryDC}
K.~Simonyan and A.~Zisserman, ``{Very Deep Convolutional Networks for Large-Scale Image Recognition},'' in \emph{International Conference on Learning Representations (ICLR)}, 2015.

\bibitem{resnet}
K.~He, X.~Zhang, S.~Ren, and J.~Sun, ``Deep residual learning for image recognition,'' in \emph{Proceedings of the IEEE conference on computer vision and pattern recognition (CVPR)}, 2016.

\bibitem{dosovitskiy2021an}
\BIBentryALTinterwordspacing
A.~Dosovitskiy, L.~Beyer, A.~Kolesnikov, D.~Weissenborn, X.~Zhai, T.~Unterthiner, M.~Dehghani, M.~Minderer, G.~Heigold, S.~Gelly, J.~Uszkoreit, and N.~Houlsby, ``{An Image is Worth 16x16 Words: Transformers for Image Recognition at Scale},'' in \emph{International Conference on Learning Representations (ICLR)}, 2021. [Online]. Available: \url{https://openreview.net/forum?id=YicbFdNTTy}
\BIBentrySTDinterwordspacing

\bibitem{Ahmed_2015_CVPR}
E.~Ahmed, M.~Jones, and T.~K. Marks, ``{An Improved Deep Learning Architecture for Person Re-Identification},'' in \emph{Proceedings of the IEEE Conference on Computer Vision and Pattern Recognition (CVPR)}, 2015.

\bibitem{Deng_2018_CVPR}
W.~Deng, L.~Zheng, Q.~Ye, G.~Kang, Y.~Yang, and J.~Jiao, ``{Image-Image Domain Adaptation With Preserved Self-Similarity and Domain-Dissimilarity for Person Re-Identification},'' in \emph{Proceedings of the IEEE Conference on Computer Vision and Pattern Recognition (CVPR)}, 2018.

\bibitem{Wang_2018_CVPR}
J.~Wang, X.~Zhu, S.~Gong, and W.~Li, ``{Transferable Joint Attribute-Identity Deep Learning for Unsupervised Person Re-Identification},'' in \emph{Proceedings of the IEEE Conference on Computer Vision and Pattern Recognition (CVPR)}, 2018.

\bibitem{re-id_uda_survey}
C.~Yang, F.~Qi, and H.~Jia, ``{Survey on Unsupervised Techniques for Person Re-Identification},'' in \emph{2nd International Conference on Computing and Data Science (CDS)}, 2021.

\bibitem{Wang2021}
D.~Wang, E.~Shelhamer, S.~Liu, B.~Olshausen, and T.~Darrell, ``{Tent: Fully Test-Time Adaptation by Entropy Minimization},'' in \emph{International Conference on Learning Representations (ICLR)}, 2021.

\bibitem{zhou2021bayesian}
A.~Zhou and S.~Levine, ``{Bayesian Adaptation for Covariate Shift},'' \emph{Advances in Neural Information Processing Systems}, 2021.

\bibitem{niu2022efficient}
S.~Niu, J.~Wu, Y.~Zhang, Y.~Chen, S.~Zheng, P.~Zhao, and M.~Tan, ``Efficient test-time model adaptation without forgetting,'' in \emph{Proceedings of the 39th International Conference on Machine Learning (ICML)}, 2022.

\bibitem{han2022generalizable}
K.~Han, C.~Si, Y.~Huang, L.~Wang, and T.~Tan, ``{Generalizable Person Re-identification via Self-Supervised Batch Norm Test-Time Adaption},'' in \emph{Proceedings of the AAAI Conference on Artificial Intelligence}, 2022.

\bibitem{Zheng_2015_ICCV}
L.~Zheng, L.~Shen, L.~Tian, S.~Wang, J.~Wang, and Q.~Tian, ``{Scalable Person Re-Identification: A Benchmark},'' in \emph{Proceedings of the IEEE International Conference on Computer Vision (ICCV)}, 2015.

\bibitem{Wei_2018_CVPR}
L.~Wei, S.~Zhang, W.~Gao, and Q.~Tian, ``{Person Transfer GAN to Bridge Domain Gap for Person Re-Identification},'' in \emph{Proceedings of the IEEE Conference on Computer Vision and Pattern Recognition (CVPR)}, 2018.

\bibitem{Sun_2019_CVPR}
X.~Sun and L.~Zheng, ``{Dissecting Person Re-Identification From the Viewpoint of Viewpoint},'' in \emph{Proceedings of the IEEE Conference on Computer Vision and Pattern Recognition (CVPR)}, 2019.

\bibitem{lin2019bottom}
Y.~Lin, X.~Dong, L.~Zheng, Y.~Yan, and Y.~Yang, ``A bottom-up clustering approach to unsupervised person re-identification,'' in \emph{Proceedings of the AAAI conference on artificial intelligence}, 2019.

\bibitem{NEURIPS2020_821fa74b}
Y.~Ge, F.~Zhu, D.~Chen, R.~Zhao, and h.~Li, ``{Self-paced Contrastive Learning with Hybrid Memory for Domain Adaptive Object Re-ID},'' in \emph{Advances in Neural Information Processing Systems}, 2020.

\bibitem{ge2020mutual}
Y.~Ge, D.~Chen, and H.~Li, ``{Mutual Mean-Teaching: Pseudo Label Refinery for Unsupervised Domain Adaptation on Person Re-identification},'' in \emph{International Conference on Learning Representations (ICLR)}, 2020.

\bibitem{zhang2021unsupervised}
M.~Zhang, K.~Liu, Y.~Li, S.~Guo, H.~Duan, Y.~Long, and Y.~Jin, ``Unsupervised domain adaptation for person re-identification via heterogeneous graph alignment,'' in \emph{Proceedings of the AAAI conference on artificial intelligence}, 2021.

\bibitem{Zhong_2019_CVPR}
Z.~Zhong, L.~Zheng, Z.~Luo, S.~Li, and Y.~Yang, ``{Invariance Matters: Exemplar Memory for Domain Adaptive Person Re-Identification},'' in \emph{Proceedings of the IEEE Conference on Computer Vision and Pattern Recognition (CVPR)}, 2019.

\bibitem{Huang_2022_CVPR}
Z.~Huang, Z.~Zhang, C.~Lan, W.~Zeng, P.~Chu, Q.~You, J.~Wang, Z.~Liu, and Z.-J. Zha, ``{Lifelong Unsupervised Domain Adaptive Person Re-Identification With Coordinated Anti-Forgetting and Adaptation},'' in \emph{Proceedings of the IEEE Conference on Computer Vision and Pattern Recognition (CVPR)}, 2022.

\bibitem{batch-norm}
S.~Ioffe and C.~Szegedy, ``Batch normalization: Accelerating deep network training by reducing internal covariate shift,'' in \emph{Proceedings of the 32nd International Conference on Machine Learning (ICML)}, 2015.

\bibitem{Benz_2021_WACV}
P.~Benz, C.~Zhang, A.~Karjauv, and I.~S. Kweon, ``{Revisiting Batch Normalization for Improving Corruption Robustness},'' in \emph{Proceedings of the IEEE Winter Conference on Applications of Computer Vision (WACV)}, 2021.

\bibitem{zhao2023delta}
\BIBentryALTinterwordspacing
B.~Zhao, C.~Chen, and S.-T. Xia, ``{DELTA}: {DEGRADATION}-{FREE} {FULLY} {TEST}-{TIME} {ADAPTATION},'' in \emph{International Conference on Learning Representations (ICLR)}, 2023. [Online]. Available: \url{https://openreview.net/forum?id=eGm22rqG93}
\BIBentrySTDinterwordspacing

\bibitem{pmlr-v80-li18a}
X.~LI, Y.~Grandvalet, and F.~Davoine, ``{Explicit Inductive Bias for Transfer Learning with Convolutional Networks},'' in \emph{Proceedings of the 35th International Conference on Machine Learning (ICML)}, 2018.

\bibitem{Qian_2019_ICCV}
Q.~Qian, L.~Shang, B.~Sun, J.~Hu, H.~Li, and R.~Jin, ``{SoftTriple Loss: Deep Metric Learning Without Triplet Sampling},'' in \emph{Proceedings of the IEEE International Conference on Computer Vision (ICCV)}, 2019.

\bibitem{adam}
D.~P. Kingma and J.~Ba, ``{Adam: A method for stochastic optimization},'' in \emph{International Conference on Learning Representations (ICLR)}, 2015.

\bibitem{openunreid}
Y.~Ge, T.~Xiao, and Z.~Zhang, ``{OpenUnReID},'' GitHub Repository, 2020, \url{https://github.com/open-mmlab/OpenUnReID}.

\bibitem{lin2021unsupervised}
X.~Lin, P.~Ren, C.-H. Yeh, L.~Yao, A.~Song, and X.~Chang, ``Unsupervised person re-identification: A systematic survey of challenges and solutions,'' \emph{arXiv preprint arXiv:2109.06057}, 2021.

\bibitem{mcinnes2018umap-software}
L.~McInnes, J.~Healy, N.~Saul, and L.~Grossberger, ``{UMAP: Uniform Manifold Approximation and Projection},'' \emph{The Journal of Open Source Software}, vol.~3, no.~29, p. 861, 2018.

\end{thebibliography}


\begin{thebibliography}{1}
\providecommand{\url}[1]{#1}
\csname url@samestyle\endcsname
\providecommand{\newblock}{\relax}
\providecommand{\bibinfo}[2]{#2}
\providecommand{\BIBentrySTDinterwordspacing}{\spaceskip=0pt\relax}
\providecommand{\BIBentryALTinterwordstretchfactor}{4}
\providecommand{\BIBentryALTinterwordspacing}{\spaceskip=\fontdimen2\font plus
\BIBentryALTinterwordstretchfactor\fontdimen3\font minus \fontdimen4\font\relax}
\providecommand{\BIBforeignlanguage}[2]{{%
\expandafter\ifx\csname l@#1\endcsname\relax
\typeout{** WARNING: IEEEtran.bst: No hyphenation pattern has been}%
\typeout{** loaded for the language `#1'. Using the pattern for}%
\typeout{** the default language instead.}%
\else
\language=\csname l@#1\endcsname
\fi
#2}}
\providecommand{\BIBdecl}{\relax}
\BIBdecl

\bibitem{Sun_2019_CVPR}
X.~Sun and L.~Zheng, ``{Dissecting Person Re-Identification From the Viewpoint of Viewpoint},'' in \emph{Proceedings of the IEEE Conference on Computer Vision and Pattern Recognition (CVPR)}, 2019.

\bibitem{Benz_2021_WACV}
P.~Benz, C.~Zhang, A.~Karjauv, and I.~S. Kweon, ``{Revisiting Batch Normalization for Improving Corruption Robustness},'' in \emph{Proceedings of the IEEE Winter Conference on Applications of Computer Vision (WACV)}, 2021.

\bibitem{han2022generalizable}
K.~Han, C.~Si, Y.~Huang, L.~Wang, and T.~Tan, ``{Generalizable Person Re-identification via Self-Supervised Batch Norm Test-Time Adaption},'' in \emph{Proceedings of the AAAI Conference on Artificial Intelligence}, 2022.

\bibitem{Wei_2018_CVPR}
L.~Wei, S.~Zhang, W.~Gao, and Q.~Tian, ``{Person Transfer GAN to Bridge Domain Gap for Person Re-Identification},'' in \emph{Proceedings of the IEEE Conference on Computer Vision and Pattern Recognition (CVPR)}, 2018.

\bibitem{mcinnes2018umap-software}
L.~McInnes, J.~Healy, N.~Saul, and L.~Grossberger, ``{UMAP: Uniform Manifold Approximation and Projection},'' \emph{The Journal of Open Source Software}, vol.~3, no.~29, p. 861, 2018.

\end{thebibliography}

\end{document}


\title{Supplementary Materials for Test-time Similarity Modification for Person Re-identification toward Temporal Distribution Shift
}

\author{\IEEEauthorblockN{Kazuki Adachi$^*${\quad}Shohei Enomoto$^*${\quad}Taku Sasaki$^*${\quad}Shin'ya Yamaguchi$^{*\dagger}$}
\IEEEauthorblockA{$^*$NTT Corporation, Tokyo, Japan \\
$^\dagger$Kyoto University, Kyoto, Japan \\
\{kazuki.adachi, shohei.enomoto, taku.sasaki, shinya.yamaguchi\}@ntt.com}
}

\maketitle

\begin{abstract}
The supplementary materials for ``Test-time Similarity Modification for Person Re-identification toward Temporal Distribution Shift.''
The detailed experimental settings (\cref{asec:exp_setting}) and additional experimental results (\cref{asec:other_result}) are provided.
\end{abstract}

\section{Experimental Settings}\label{asec:exp_setting}

\subsection{Evaluation Metric}\label{assec:evaluation_metric}
We used the top-1 cumulative matching characteristics (CMC) accuracy as the evaluation metric.
CMC is one of the most popular evaluation metrics in re-id.
Top-1 CMC accuracy measures the ratio of the query images for which the nearest gallery feature has the same identity as the query feature:
\begin{equation}
\text{CMC}_1 = \frac{1}{n^\text{q}} \sum_{i=1}^{n^\text{q}} \indicator \left[ y(\mathcal{N}^\text{g}(\mathbf{z}^\text{q}_i))=y_i \right],
\end{equation}
where $n^\text{q}$ is the total number of queries, $\mathbf{z}^\text{q}_i$ is a query feature, $y_i$ is the ground-truth of the person identity of $\mathbf{z}^\text{q}_i$, $\mathcal{N}^\text{g}(\cdot)$ is the nearest gallery feature, $y(\cdot)$ is the person identity label of a given gallery feature, and $\indicator[\cdot]$ is the indicator function. 

\subsection{Corruption}\label{assec:condition_change}
In the corruption scenario, we corrupted the query images by varying their brightness, adding Gaussian blur, and pixelate.
Here, we describe these corruptions.

\textbf{Brightness} multiplies a factor $s$ to each pixel value of the images.
The pixel values are clipped when they exceed the range of $[0,255]$.
In the experiment, we changed $s$ as follows: [0.75, 0.5, 0.75,$ $ 1, 1.25, 1.5, 1.25, 1].

\textbf{Gaussian blur} convolves a Gaussian filter with the images.
We changed the filter size $k$ as follows: [7, 15, 29, 35, 43].
The standard deviation of the Gaussian filter was automatically determined based on $k$.
For more details, please refer to torchvision's documentation~(torchvision.transforms.functional.gaussian\_blur).

\textbf{Pixelate} reduces the pixel information by shrinking and then expanding the images.
We resized the images, first by a scale factor $s ~(<1)$, then back to the original size.
We changed the scale factor $s$ as follows: [0.6, 0.5, 0.4, 0.3, 0.25].

\begin{figure*}[tb]
\centering
\begin{tabular}{ccccc}
\includegraphics[alt={Original image},width=0.16\linewidth]{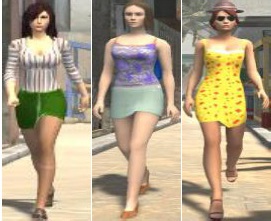} &
\includegraphics[alt={Brightness at s=0.25},width=0.16\linewidth]{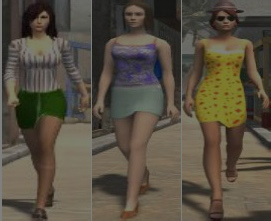} &
\includegraphics[alt={Brightness at s=1.5},width=0.16\linewidth]{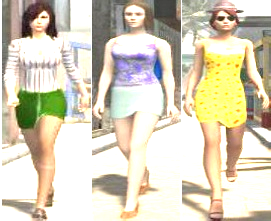} &
\includegraphics[alt={Gaussian blur at k=43},width=0.16\linewidth]{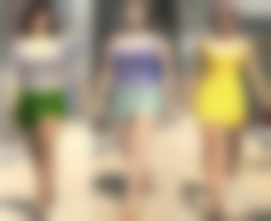} &
\includegraphics[alt={Pixelate at s=0.25},width=0.16\linewidth]{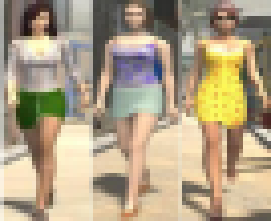} \\
(a) & (b) & (c) & (d) & (e)
\end{tabular}
\caption{Examples of the corruptions.
(a) Original image.
(b) Brightness at $s=0.25$.
(c) Brightness at $s=1.5$.
(d) Gaussian blur at $k=43$.
(e) Pixelate at $s=0.25$.
These images are sampled from PersonX~\cite{Sun_2019_CVPR}.}
\label{afig:sample_images}
\end{figure*}

\cref{afig:sample_images} illustrates examples of the corruptions.

\section{Other Results}\label{asec:other_result}

\begin{table}[tb]
\centering
\caption{Average top-1 CMC accuracy \vs~$\lambda$ in the corruption scenario.}
\label{atab:corruption_anti-forgetting}
\setlength{\tabcolsep}{3pt}
\begin{tabular}{lllll}\toprule
~ & ~ & \multicolumn{3}{c}{$\lambda$} \\
Corruption & Source & 0.0 & 0.0001 & 0.001 \\ \midrule
\multirow{3}{*}{Brightness} & MA & $90.74_{\pm 0.06}$ & $90.62_{\pm 0.03}$ & $90.40_{\pm 0.05}$ \\
~ & MS & $66.16_{\pm 0.03}$ & $65.64_{\pm 0.04}$ & $65.40_{\pm 0.02}$ \\
~ & PX & $91.91_{\pm 0.02}$ & $91.26_{\pm 0.02}$ & $90.24_{\pm 0.04}$ \\ \midrule
\multirow{3}{*}{Gaussian-blur} & MA & $87.81_{\pm 0.10}$ & $87.50_{\pm 0.07}$ & $86.16_{\pm 0.05}$ \\
~ & MS & $55.36_{\pm 0.04}$ & $53.25_{\pm 0.06}$ & $44.23_{\pm 0.94}$ \\
~ & PX & $90.97_{\pm 0.02}$ & $90.35_{\pm 0.04}$ & $88.79_{\pm 0.06}$ \\ \midrule
\multirow{3}{*}{Pixelate} & MA & $90.55_{\pm 0.08}$ & $90.50_{\pm 0.05}$ & $90.18_{\pm 0.06}$ \\
~ & MS & $65.01_{\pm 0.02}$ & $64.53_{\pm 0.04}$ & $64.02_{\pm 0.06}$ \\
~ & PX & $91.87_{\pm 0.02}$ & $91.38_{\pm 0.05}$ & $90.45_{\pm 0.03}$ \\ \bottomrule
\end{tabular}
\end{table}

\begin{table}[tb]
\centering
\caption{Average top-1 CMC accuracy \vs~$\lambda$ in the location change scenario.}
\label{atab:location_anti-forgetting}
{\setlength{\tabcolsep}{3pt}
\begin{tabular}{lllll}\toprule
~ & ~ & \multicolumn{3}{c}{Phase} \\
Source & $\lambda$ & 1 & 2 & 3  \\ \midrule
\multirow{4}{*}{Market-1501} & ~ & PX & MS & MA \\
~ & 0.0 & $69.04_{\pm 0.28}$ & $33.05_{\pm 0.22}$ & $88.59_{\pm 0.23}$ \\
~ & 0.0001 & $68.30_{\pm 0.35}$ & $32.89_{\pm 0.30}$ & $89.33_{\pm 0.16}$ \\
~ & 0.001 & $65.91_{\pm 0.16}$ & $32.04_{\pm 0.18}$ & $89.93_{\pm 0.23}$ \\ \midrule
\multirow{4}{*}{MSMT17} & ~ & MA & PX & MS \\
~ & 0.0 & $63.70_{\pm 0.26}$ & $63.06_{\pm 0.11}$ & $63.18_{\pm 0.16}$ \\
~ & 0.0001 & $63.62_{\pm 0.24}$ & $61.93_{\pm 0.25}$ & $64.47_{\pm 0.17}$ \\
~ & 0.001 & $62.68_{\pm 0.19}$ & $58.93_{\pm 0.26}$ & $65.18_{\pm 0.03}$ \\ \midrule
\multirow{4}{*}{PersonX} & ~ & MS & MA & PX \\
~ & 0.0 & $25.72_{\pm 0.20}$ & $55.54_{\pm 0.69}$ & $90.58_{\pm 0.20}$ \\
~ & 0.0001 & $25.34_{\pm 0.19}$ & $54.55_{\pm 0.26}$ & $90.69_{\pm 0.28}$ \\
~ & 0.001 & $24.79_{\pm 0.12}$ & $52.35_{\pm 0.36}$ & $90.12_{\pm 0.08}$ \\ \bottomrule
\end{tabular}
}
\end{table}

\inputfig{1.0}{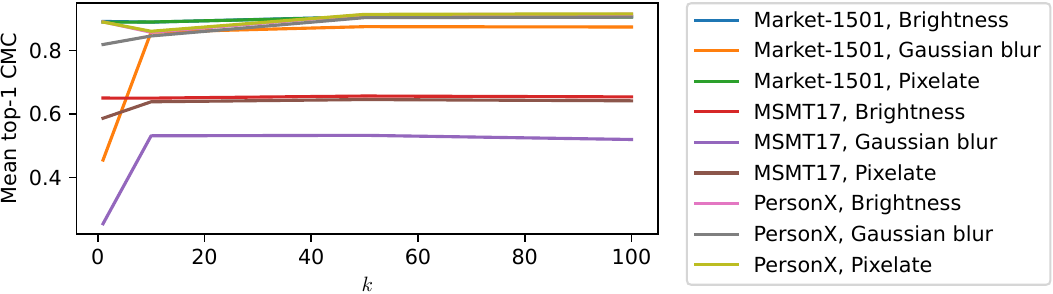}{Top-1 CMC accuracy {\vs} $k$ in the corruption scenario.}{afig:corruption_top-k}{Top-1 CMC accuracy vs. k in the corruption scenario.}

\subsection{Effect of Batch Size}\label{assec:batch-size}
\begin{table*}[tb]
\centering
\caption{Average top-1 CMC accuracy \vs~batch size in the corruption scenario.
Each CMC accuracy is the mean over the phases.}
\label{atab:cmc_corruption}
{\footnotesize
\setlength{\tabcolsep}{4pt}
\begin{tabular}{lllllllll} \toprule
~ & ~ & ~ & \multicolumn{6}{c}{Batch size} \\
Corruption & Source & Method & 64 & 32 & 16 & 8 & 2 & 1 \\ \midrule
\multirow{15}{*}{Brightness} & \multirow{5}{*}{MA} & No-adapt & $89.11_{\pm 0.00}$ & $89.14_{\pm 0.00}$ & $89.14_{\pm 0.00}$ & $89.14_{\pm 0.00}$ & $89.14_{\pm 0.00}$ & $89.14_{\pm 0.00}$ \\
~ & ~ & BN-adapt~\cite{Benz_2021_WACV} & $89.17_{\pm 0.08}$ & $88.91_{\pm 0.10}$ & $88.45_{\pm 0.09}$ & $87.54_{\pm 0.09}$ & $75.88_{\pm 0.19}$ & - \\
~ & ~ & SourceTent & $80.52_{\pm 0.26}$ & $75.36_{\pm 0.29}$ & $63.45_{\pm 0.50}$ & $38.52_{\pm 2.58}$ & $9.17_{\pm 0.48}$ & - \\
~ & ~ & BNTA~\cite{han2022generalizable} & $1.42_{\pm 0.08}$ & $0.91_{\pm 0.03}$ & $0.46_{\pm 0.02}$ & $0.54_{\pm 0.05}$ & - & - \\
~ & ~ & TEMP & $\mathbf{90.58_{\pm 0.03}}$ & $\mathbf{90.63_{\pm 0.04}}$ & $\mathbf{90.62_{\pm 0.03}}$ & $\mathbf{90.59_{\pm 0.08}}$ & $\mathbf{90.44_{\pm 0.04}}$ & $\mathbf{90.64_{\pm 0.08}}$ \\ \cmidrule{2-9}
~ & \multirow{5}{*}{MS} & No-adapt & $65.00_{\pm 0.00}$ & $65.00_{\pm 0.00}$ & $65.01_{\pm 0.00}$ & $64.99_{\pm 0.00}$ & $65.00_{\pm 0.00}$ & $65.00_{\pm 0.00}$ \\
~ & ~ & BN-adapt~\cite{Benz_2021_WACV} & $64.57_{\pm 0.05}$ & $64.18_{\pm 0.01}$ & $63.32_{\pm 0.05}$ & $61.66_{\pm 0.03}$ & $44.70_{\pm 0.11}$ & - \\
~ & ~ & SourceTent & $39.87_{\pm 1.02}$ & $26.18_{\pm 0.68}$ & $17.71_{\pm 0.72}$ & $10.17_{\pm 0.52}$ & $3.32_{\pm 0.42}$ & - \\
~ & ~ & BNTA~\cite{han2022generalizable} & $0.56_{\pm 0.10}$ & $0.30_{\pm 0.01}$ & $0.19_{\pm 0.03}$ & $0.15_{\pm 0.01}$ & - & - \\
~ & ~ & TEMP & $\mathbf{65.74_{\pm 0.00}}$ & $\mathbf{65.75_{\pm 0.02}}$ & $\mathbf{65.64_{\pm 0.04}}$ & $\mathbf{65.52_{\pm 0.05}}$ & $\mathbf{64.96_{\pm 0.08}}$ & $\mathbf{65.80_{\pm 0.02}}$ \\ \cmidrule{2-9}
~ & \multirow{5}{*}{PX} & No-adapt & $89.02_{\pm 0.00}$ & $89.02_{\pm 0.00}$ & $89.00_{\pm 0.00}$ & $89.00_{\pm 0.00}$ & $89.01_{\pm 0.00}$ & $89.01_{\pm 0.00}$ \\
~ & ~ & BN-adapt~\cite{Benz_2021_WACV} & $87.12_{\pm 0.03}$ & $86.95_{\pm 0.02}$ & $86.53_{\pm 0.04}$ & $85.72_{\pm 0.05}$ & $73.00_{\pm 0.11}$ & - \\
~ & ~ & SourceTent & $86.42_{\pm 0.13}$ & $81.47_{\pm 1.00}$ & $68.81_{\pm 1.05}$ & $51.55_{\pm 2.62}$ & $13.16_{\pm 0.79}$ & - \\
~ & ~ & BNTA~\cite{han2022generalizable} & $1.87_{\pm 0.12}$ & $1.58_{\pm 0.03}$ & $0.75_{\pm 0.06}$ & $1.07_{\pm 0.26}$ & - & - \\
~ & ~ & TEMP & $\mathbf{91.18_{\pm 0.05}}$ & $\mathbf{91.24_{\pm 0.02}}$ & $\mathbf{91.26_{\pm 0.02}}$ & $\mathbf{91.18_{\pm 0.05}}$ & $\mathbf{91.00_{\pm 0.09}}$ & $\mathbf{91.15_{\pm 0.01}}$ \\ \midrule
\multirow{15}{*}{Gaussian-blur} & \multirow{5}{*}{MA} & No-adapt & $45.47_{\pm 0.00}$ & $45.49_{\pm 0.00}$ & $45.49_{\pm 0.00}$ & $45.48_{\pm 0.00}$ & $45.48_{\pm 0.00}$ & $45.47_{\pm 0.00}$ \\
~ & ~ & BN-adapt~\cite{Benz_2021_WACV} & $78.81_{\pm 0.06}$ & $78.45_{\pm 0.02}$ & $77.72_{\pm 0.12}$ & $76.20_{\pm 0.15}$ & $61.33_{\pm 0.38}$ & - \\
~ & ~ & SourceTent & $65.45_{\pm 0.14}$ & $61.16_{\pm 0.15}$ & $54.18_{\pm 0.23}$ & $42.01_{\pm 0.91}$ & $13.83_{\pm 0.22}$ & - \\
~ & ~ & BNTA~\cite{han2022generalizable} & $2.17_{\pm 0.06}$ & $1.35_{\pm 0.19}$ & $0.64_{\pm 0.02}$ & $0.75_{\pm 0.13}$ & - & - \\
~ & ~ & TEMP & $\mathbf{86.40_{\pm 0.02}}$ & $\mathbf{87.17_{\pm 0.07}}$ & $\mathbf{87.50_{\pm 0.07}}$ & $\mathbf{87.68_{\pm 0.02}}$ & $\mathbf{87.63_{\pm 0.12}}$ & $\mathbf{86.84_{\pm 0.07}}$ \\ \cmidrule{2-9}
~ & \multirow{5}{*}{MS} & No-adapt & $25.38_{\pm 0.00}$ & $25.38_{\pm 0.00}$ & $25.38_{\pm 0.00}$ & $25.38_{\pm 0.00}$ & $25.37_{\pm 0.00}$ & $25.36_{\pm 0.00}$ \\
~ & ~ & BN-adapt~\cite{Benz_2021_WACV} & $41.57_{\pm 0.09}$ & $41.12_{\pm 0.04}$ & $40.09_{\pm 0.09}$ & $38.31_{\pm 0.05}$ & $23.92_{\pm 0.08}$ & - \\
~ & ~ & SourceTent & $27.94_{\pm 0.22}$ & $24.02_{\pm 0.34}$ & $19.14_{\pm 0.91}$ & $12.94_{\pm 0.37}$ & $4.82_{\pm 0.12}$ & - \\
~ & ~ & BNTA~\cite{han2022generalizable} & $0.67_{\pm 0.04}$ & $0.39_{\pm 0.02}$ & $0.28_{\pm 0.08}$ & $0.20_{\pm 0.03}$ & - & - \\
~ & ~ & TEMP & $\mathbf{52.91_{\pm 0.08}}$ & $\mathbf{53.15_{\pm 0.09}}$ & $\mathbf{53.25_{\pm 0.06}}$ & $\mathbf{53.20_{\pm 0.04}}$ & $\mathbf{51.90_{\pm 0.03}}$ & $\mathbf{52.97_{\pm 0.05}}$ \\ \cmidrule{2-9}
~ & \multirow{5}{*}{PX} & No-adapt & $81.89_{\pm 0.00}$ & $81.89_{\pm 0.00}$ & $81.87_{\pm 0.00}$ & $81.87_{\pm 0.00}$ & $81.88_{\pm 0.00}$ & $81.88_{\pm 0.00}$ \\
~ & ~ & BN-adapt~\cite{Benz_2021_WACV} & $84.68_{\pm 0.14}$ & $84.36_{\pm 0.12}$ & $83.79_{\pm 0.07}$ & $82.37_{\pm 0.08}$ & $66.41_{\pm 0.09}$ & - \\
~ & ~ & SourceTent & $79.79_{\pm 0.30}$ & $75.85_{\pm 0.87}$ & $68.10_{\pm 1.36}$ & $55.43_{\pm 1.70}$ & $19.00_{\pm 0.67}$ & - \\
~ & ~ & BNTA~\cite{han2022generalizable} & $2.37_{\pm 0.06}$ & $1.83_{\pm 0.22}$ & $1.23_{\pm 0.08}$ & $1.26_{\pm 0.22}$ & - & - \\
~ & ~ & TEMP & $\mathbf{90.05_{\pm 0.01}}$ & $\mathbf{90.20_{\pm 0.05}}$ & $\mathbf{90.35_{\pm 0.04}}$ & $\mathbf{90.32_{\pm 0.07}}$ & $\mathbf{88.06_{\pm 2.96}}$ & $\mathbf{89.96_{\pm 0.04}}$ \\ \midrule
\multirow{15}{*}{Pixelate} & \multirow{5}{*}{MA} & No-adapt & $88.91_{\pm 0.00}$ & $88.92_{\pm 0.00}$ & $88.92_{\pm 0.00}$ & $88.92_{\pm 0.00}$ & $88.91_{\pm 0.00}$ & $88.91_{\pm 0.00}$ \\
~ & ~ & BN-adapt~\cite{Benz_2021_WACV} & $88.89_{\pm 0.13}$ & $88.56_{\pm 0.05}$ & $88.10_{\pm 0.12}$ & $86.99_{\pm 0.16}$ & $75.21_{\pm 0.07}$ & - \\
~ & ~ & SourceTent & $85.69_{\pm 0.18}$ & $82.74_{\pm 0.24}$ & $75.81_{\pm 0.36}$ & $57.07_{\pm 1.96}$ & $14.76_{\pm 0.36}$ & - \\
~ & ~ & BNTA~\cite{han2022generalizable} & $2.25_{\pm 0.08}$ & $1.30_{\pm 0.03}$ & $0.83_{\pm 0.07}$ & $0.76_{\pm 0.06}$ & - & - \\
~ & ~ & TEMP & $\mathbf{90.49_{\pm 0.03}}$ & $\mathbf{90.50_{\pm 0.04}}$ & $\mathbf{90.50_{\pm 0.05}}$ & $\mathbf{90.49_{\pm 0.05}}$ & $\mathbf{90.30_{\pm 0.09}}$ & $\mathbf{90.48_{\pm 0.06}}$ \\ \cmidrule{2-9}
~ & \multirow{5}{*}{MS} & No-adapt & $58.63_{\pm 0.00}$ & $58.62_{\pm 0.00}$ & $58.61_{\pm 0.00}$ & $58.62_{\pm 0.00}$ & $58.63_{\pm 0.00}$ & $58.64_{\pm 0.00}$ \\
~ & ~ & BN-adapt~\cite{Benz_2021_WACV} & $62.18_{\pm 0.08}$ & $61.76_{\pm 0.05}$ & $60.82_{\pm 0.10}$ & $59.03_{\pm 0.03}$ & $40.94_{\pm 0.10}$ & - \\
~ & ~ & SourceTent & $48.70_{\pm 0.44}$ & $37.13_{\pm 0.83}$ & $25.31_{\pm 1.00}$ & $15.20_{\pm 0.84}$ & $5.38_{\pm 0.65}$ & - \\
~ & ~ & BNTA~\cite{han2022generalizable} & $0.78_{\pm 0.06}$ & $0.61_{\pm 0.29}$ & $0.60_{\pm 0.36}$ & $0.36_{\pm 0.28}$ & - & - \\
~ & ~ & TEMP & $\mathbf{64.67_{\pm 0.01}}$ & $\mathbf{64.64_{\pm 0.04}}$ & $\mathbf{64.53_{\pm 0.04}}$ & $\mathbf{64.37_{\pm 0.03}}$ & $\mathbf{63.70_{\pm 0.06}}$ & $\mathbf{64.74_{\pm 0.03}}$ \\ \cmidrule{2-9}
~ & \multirow{5}{*}{PX} & No-adapt & $88.96_{\pm 0.00}$ & $88.96_{\pm 0.00}$ & $88.96_{\pm 0.00}$ & $88.96_{\pm 0.00}$ & $88.96_{\pm 0.00}$ & $88.96_{\pm 0.00}$ \\
~ & ~ & BN-adapt~\cite{Benz_2021_WACV} & $87.30_{\pm 0.05}$ & $87.15_{\pm 0.02}$ & $86.81_{\pm 0.12}$ & $85.99_{\pm 0.07}$ & $72.92_{\pm 0.13}$ & - \\
~ & ~ & SourceTent & $88.72_{\pm 0.18}$ & $86.41_{\pm 0.55}$ & $80.94_{\pm 0.65}$ & $68.38_{\pm 2.19}$ & $22.00_{\pm 1.08}$ & - \\
~ & ~ & BNTA~\cite{han2022generalizable} & $2.34_{\pm 0.11}$ & $2.37_{\pm 0.52}$ & $1.29_{\pm 0.23}$ & $1.05_{\pm 0.07}$ & - & - \\
~ & ~ & TEMP & $\mathbf{91.24_{\pm 0.03}}$ & $\mathbf{91.30_{\pm 0.02}}$ & $\mathbf{91.38_{\pm 0.05}}$ & $\mathbf{91.31_{\pm 0.05}}$ & $\mathbf{91.05_{\pm 0.15}}$ & $\mathbf{91.26_{\pm 0.01}}$ \\ \bottomrule
\end{tabular}
}
\end{table*}

In the online setting, models often have to predict incoming data instantly rather than waiting to accumulate sufficient data for making a batch.
We ran TTA with batch sizes of $\{64,32,16,8,2,1 \}$.

\textbf{Location change}:  Please refer to the main paper.

\textbf{Image Corruption}: \cref{atab:cmc_corruption} shows the effect of the batch size tested in the corruption case.
Similar to the location change scenario, {\proposedmethod} consistently outperformed the baselines in almost all cases while BN-adapt had low accuracy with a small batch size.

\subsection{Additional Results in the Corruption Scenario}\label{assec:accuracy-transition}
\begin{figure}[tb]
\centering
\begin{tabular}{c}
\includegraphics[alt={Top-1 CMC accuracy vs. iteration in the image corruption scenario (brightness)},width=0.95\linewidth]{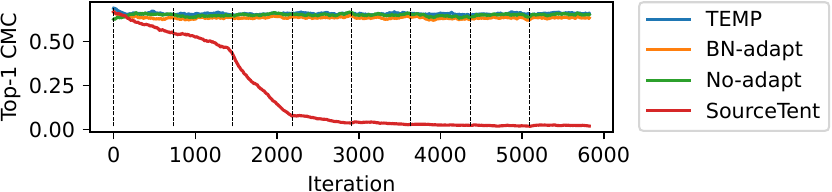} \\
\includegraphics[alt={Top-1 CMC accuracy vs. iteration in the image corruption scenario (Gaussian blur)},width=0.95\linewidth]{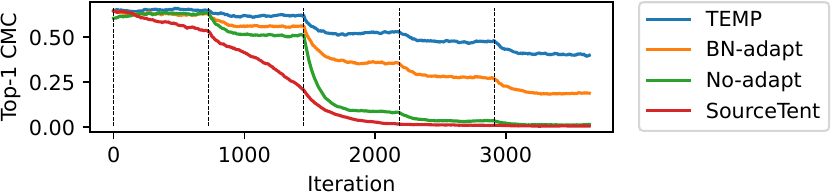} \\
\includegraphics[alt={Top-1 CMC accuracy vs. iteration in the image corruption scenario (pixelate)},width=0.95\linewidth]{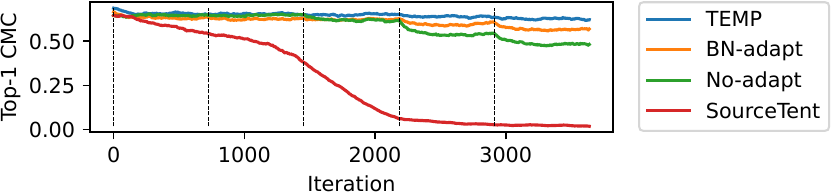}
\end{tabular}
\caption{Top-1 CMC accuracy {\vs} iteration in the image corruption scenario.
The source dataset is MSMT17.
The top, middle, and bottom rows represent brightness, Gaussian blur, and pixelate corruptions, respectively.
The vertical dashed lines represent the iterations when the strengths of the corruptions change.
We plot the exponential moving average of the CMCs computed for each mini-batch.}
\label{afig:cmc_corruption_msmt17}
\end{figure}

\begin{figure}[tb]
\centering
\begin{tabular}{c}
\includegraphics[alt={Top-1 CMC accuracy vs. iteration in the image corruption scenario (brightness)},width=0.95\linewidth]{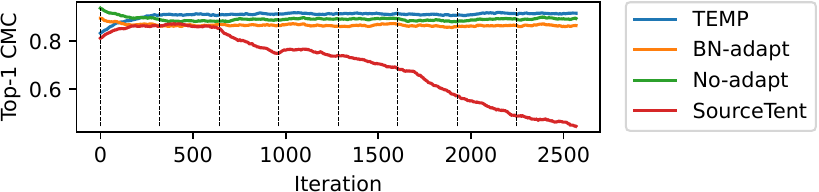} \\
\includegraphics[alt={Top-1 CMC accuracy vs. iteration in the image corruption scenario (Gaussian blur)},width=0.95\linewidth]{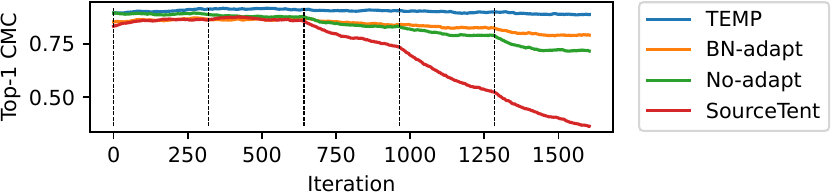} \\
\includegraphics[alt={Top-1 CMC accuracy vs. iteration in the image corruption scenario (pixelate)},width=0.95\linewidth]{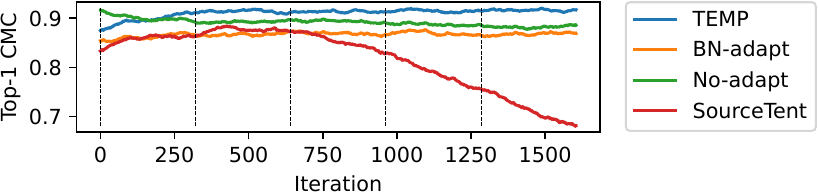}
\end{tabular}
\caption{Top-1 CMC accuracy {\vs} iteration in the image corruption scenario.
The source dataset is PersonX.
The top, middle, and bottom rows represent brightness, Gaussian blur, and pixelate, respectively.
The vertical dashed lines represent the iterations when the strengths of the corruptions change.
We plot the exponential moving average of the CMCs computed for each mini-batch.}
\label{afig:cmc_corruption_personx}
\end{figure}

\cref{afig:cmc_corruption_msmt17,afig:cmc_corruption_personx} show top-1 CMC accuracies for each corruption where the source dataset is MSMT17~\cite{Wei_2018_CVPR} and PersonX~\cite{Sun_2019_CVPR}.
{\proposedmethod} has the best top-1 CMC accuracy in most cases.

For the location change scenario, please refer to the main paper.

\subsection{Effect of k in the Corruption Scenario}

We examined the effect of varying $k$ in the corruption scenario.
\cref{afig:corruption_top-k} shows the results.
Similar to the location change scenario shown in the main paper, about $k=50$ consistently produces better results.

For the location change scenario, please refer to the main paper.

\subsection{Effect of the Anti-forgetting Regularization}\label{assec:ablation}

We investigated the effect of the anti-forgetting regularization.
We varied the weight of the regularization $\lambda$ within $\{ 0, 0.0001, 0.001 \}$ and ran {\proposedmethod}.
\cref{atab:location_anti-forgetting,atab:corruption_anti-forgetting} show the average top-1 CMC accuracy in the location change and corruption scenarios.
In \cref{atab:location_anti-forgetting}, a larger $\lambda$ improves accuracy when the distribution returns to the source one (phase 3) since the anti-forgetting regularization prevents the model parameters from changing drastically.
On the other hand, the accuracy slightly degrades when the distribution differs from the source one (phase 1,2 in \cref{atab:location_anti-forgetting} and the corruption scenario in \cref{atab:corruption_anti-forgetting}).
Thus, there is a trade-off between source performance and adaptability.

\begin{figure*}[tb]
\centering
\begin{tabular}{cc}
\includegraphics[alt={Comparison of gallery feature selection strategies (location change).},height=10em]{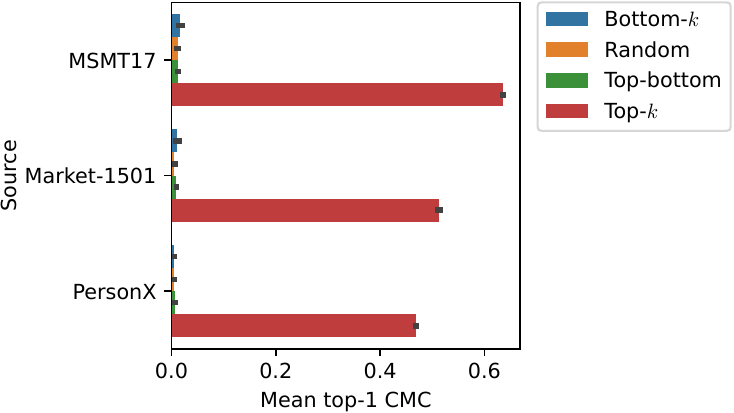}
&
\includegraphics[alt={Comparison of gallery feature selection strategies (corruption).},height=10em]{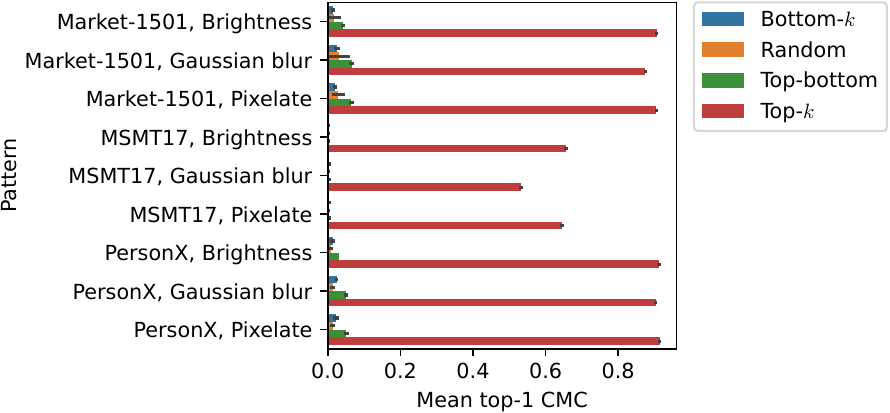}
\\
Location change & Corruption
\end{tabular}
\caption{Comparison of gallery feature selection strategies.
Top-$k$ had the highest accuracy.}
\label{afig:selection-type}
\end{figure*}

\subsection{Effect of the Similarity Modification}\label{assec:feature_umap}
\begin{figure}[tb]
\centering
\begin{tabular}{cc}
\includegraphics[alt={UMAP visualization of the query and gallery features (No-adapt)},width=0.45\linewidth]{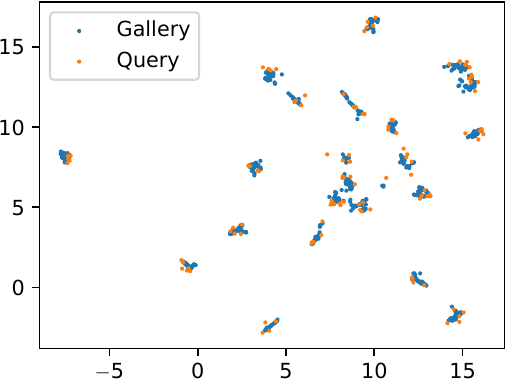}
&
\includegraphics[alt={UMAP visualization of the query and gallery features ({\proposedmethod})},width=0.45\linewidth]{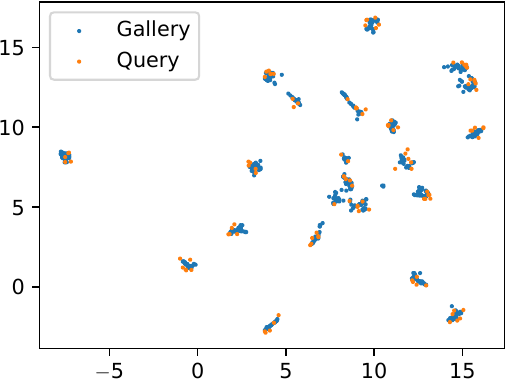}
\\
No-adapt & {\proposedmethod}
\end{tabular}
\caption{UMAP~\cite{mcinnes2018umap-software} visualization of the query and gallery features in the location change scenario (Market-1501 $\rightarrow$ PersonX).}
\label{afig:umap_location-change}
\end{figure}

Similarly to what was described in Sec~5.3.5 of the main paper, we visualized the query and gallery features with UMAP~\cite{mcinnes2018umap-software}.
We extracted the gallery features of PersonX with the feature extractor trained on Market-1501.
Then, we ran TTA and extracted the query features with the adapted model.
\cref{afig:umap_location-change} shows the UMAP visualization.
Compared with No-adapt, {\proposedmethod} reduces the number of query features that are isolated from the gallery clusters.

\subsection{Gallery Feature Selection Strategy}\label{assec:selection_strategy}

\inputfig{1.0}{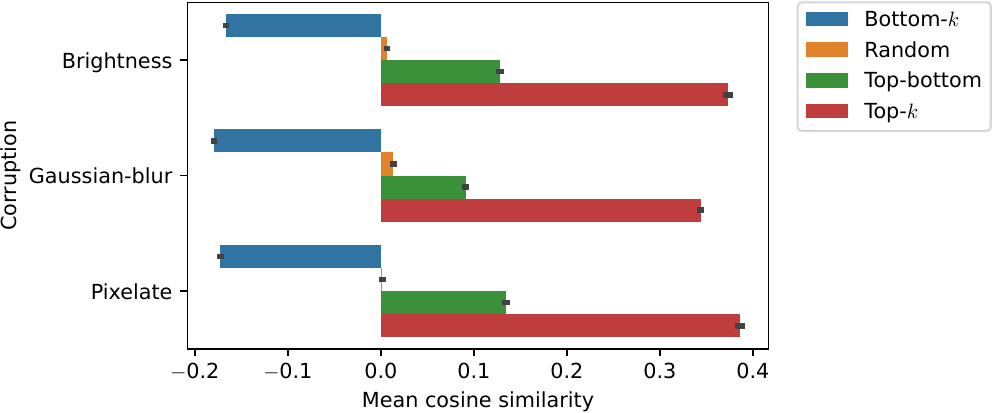}{Mean cosine similarity of the query features and the selected $k$ gallery features in each selection strategy.
The source model is trained on Market-1501.}{afig:feature_similarity}{Mean cosine similarity of the query features and the selected k gallery features in each selection strategy.}

For computing the probability $\hat{p}_{ij}$ from the feature similarity in Eq.~(3) of the main paper, we selected the top-$k$ similar gallery features in terms of the similarity $s_{ij}$.
We examined other selection strategies to investigate a better one:

\textbf{Top-$k$}: selects top-$k$ similar features as in the main paper. 

\textbf{Bottom-$k$}: selects top-$k$ \emph{dissimilar} features.

\textbf{Top-bottom}: selects top-$k/2$ similar and top-$k/2$ dissimilar features.

\textbf{Random}: randomly selects $k$ features.

\cref{afig:selection-type} shows the results.
Only the top-$k$ worked because modifying neighbors of the query features in the feature space is important.
On the other hand, using dissimilar gallery features that do not affect re-id inference negatively affected.
\cref{afig:feature_similarity} shows the mean cosine similarities between the query features and the selected $k$ gallery features.
Although we expected the other strategies to make irrelevant gallery features distant from the query one, such an effect can occur within the top-$k$ similar gallery features since we used softmax for computing $\hat{p}_{ij}$.

\bibliographystyle{IEEEtran}
\bibliography{supplement}